\documentclass[lettersize,journal]{IEEEtran}
\usepackage{amsmath,amsfonts}
\usepackage{algorithmic}
\usepackage{algorithm}
\usepackage{enumitem}
\usepackage{array}
\usepackage{textcomp}
\usepackage{stfloats}
\usepackage{url}
\usepackage{verbatim}
\usepackage{graphicx}
\usepackage{cite}
\usepackage{times}
\usepackage{soul}
\usepackage{amsmath}
\usepackage{amssymb}
\usepackage{amsthm}
\usepackage{booktabs}
\usepackage{algorithm}
\usepackage{multirow}
\usepackage{subfigure}
\usepackage{tabularx}
\usepackage{subcaption}
\usepackage[switch]{lineno}

\usepackage{xcolor}

\begin{document}

\title{Fairness in Augmented Graph Learning: A Survey}

\author{Renqiang Luo,
        Huafei Huang,
        Ziqi Xu,
        Xikun Zhang,
        Enyan Dai,
        Bo Yang,
        Feng Xia,~\IEEEmembership{IEEE Fellow}
\thanks{Renqiang Luo and Bo Yang are with the College of Computer Science and Technology, Jilin University, Changchun 130012, China (e-mail: \{lrenqiang,ybo\}@jlu.edu.cn).}
\thanks{Huafei Huang is with the School of Computer Science and Information Technology, Adelaide University, Adelaide, SA 5095, Australia (e-mail: hhuafei@outlook.com).}
\thanks{Ziqi Xu, Xikun Zhang and Feng Xia are with the School of Computing Technologies, RMIT University, Melbourne, VIC 3000, Australia (e-mail: \{ziqi.xu, xikun.zhang\}@rmit.edu.au, f.xia@ieee.org).}
\thanks{Enyan Dai is with Hong Kong University of Science and Technology (Guangzhou), Guangzhou, China (e-mail: enyandai@hkust-gz.edu.cn)}
\thanks{Corresponding author: Bo Yang.}}

\markboth{IEEE Transactions on Knowledge and Data Engineering,~Vol.~0, No.~0, March~2026}%
{Shell \MakeLowercase{\textit{et al.}}: A Sample Article Using IEEEtran.cls for IEEE Journals}


\maketitle

\begin{abstract}
Graph learning has evolved into Augmented Graph Learning (AGL) by integrating specialized machine learning (ML) techniques. 
Examples include federated learning, graph transformers, and graph condensation.
While enhancing model utility, AGL introduces unique intersectional fairness challenges that traditional GNN debiasing frameworks, which primarily focus on message-passing regulations, fail to address. 
This paper provides a systematic investigation into this emerging field, termed FairGX.
We first delineate the shift from conventional fairness-aware graph learning to the FairGX paradigm, identifying novel bias sources inherent in ML augmentations, such as dual-side disparities in federated aggregation and attention-head skewness. 
A structured taxonomy is established to categorize existing literature based on their technical integration and fairness objectives. 
Furthermore, we analyze the impact of diverse ML paradigms on algorithmic equity, emphasizing the unique challenges in human-centered applications and the absence of a unified framework. 
We conclude by identifying five critical future directions, including novel metrics for AGL, fairness-privacy synergy, and Fairness-aware LLM4Graph/Graph4LLM. 
This survey serves as a foundational roadmap for developing robust and equitable graph systems in complex ML environments. 
\end{abstract}


\section{Introduction}
\IEEEPARstart{G}{raph} learning has demonstrated remarkable effectiveness in handling graph-structured data, driving advancements across various real-world applications~\cite{xia2021graph}. 
Despite its success, traditional graph learning methods face inherent challenges, including over-smoothing~\cite{guo2023contranorm}, over-squashing~\cite{he2023generalization}, and high computational complexity~\cite{gao2024graph}. 
Additionally, practical applications often demand capabilities such as privacy protection~\cite{tang2024personlized}, adaptability to evolving data~\cite{pei2024memory}, and scalability for large-scale graphs~\cite{traag2019louvain}. 
To address these issues, specialised machine learning (ML) techniques have been introduced as augmentations, forming the framework of \textbf{A}ugmented \textbf{G}raph \textbf{L}earning (AGL). 
For example, graph federated learning and dynamic graph learning address data privacy and evolving structures, respectively~\cite{wan2024federated, chen2024easydgl}. 
Furthermore, graph transformers mitigate over-smoothing~\cite{wu2023demystifying}, while graph condensation and partitioning methods optimize efficiency for processing large graphs~\cite{jin2022graph, modak2024cpa}.
Overall, these advancements enable AGL to overcome the limitations of traditional graph learning and meet the demands of modern applications.

\begin{figure}[t] 
    \centering
	\includegraphics[width=0.48\textwidth]{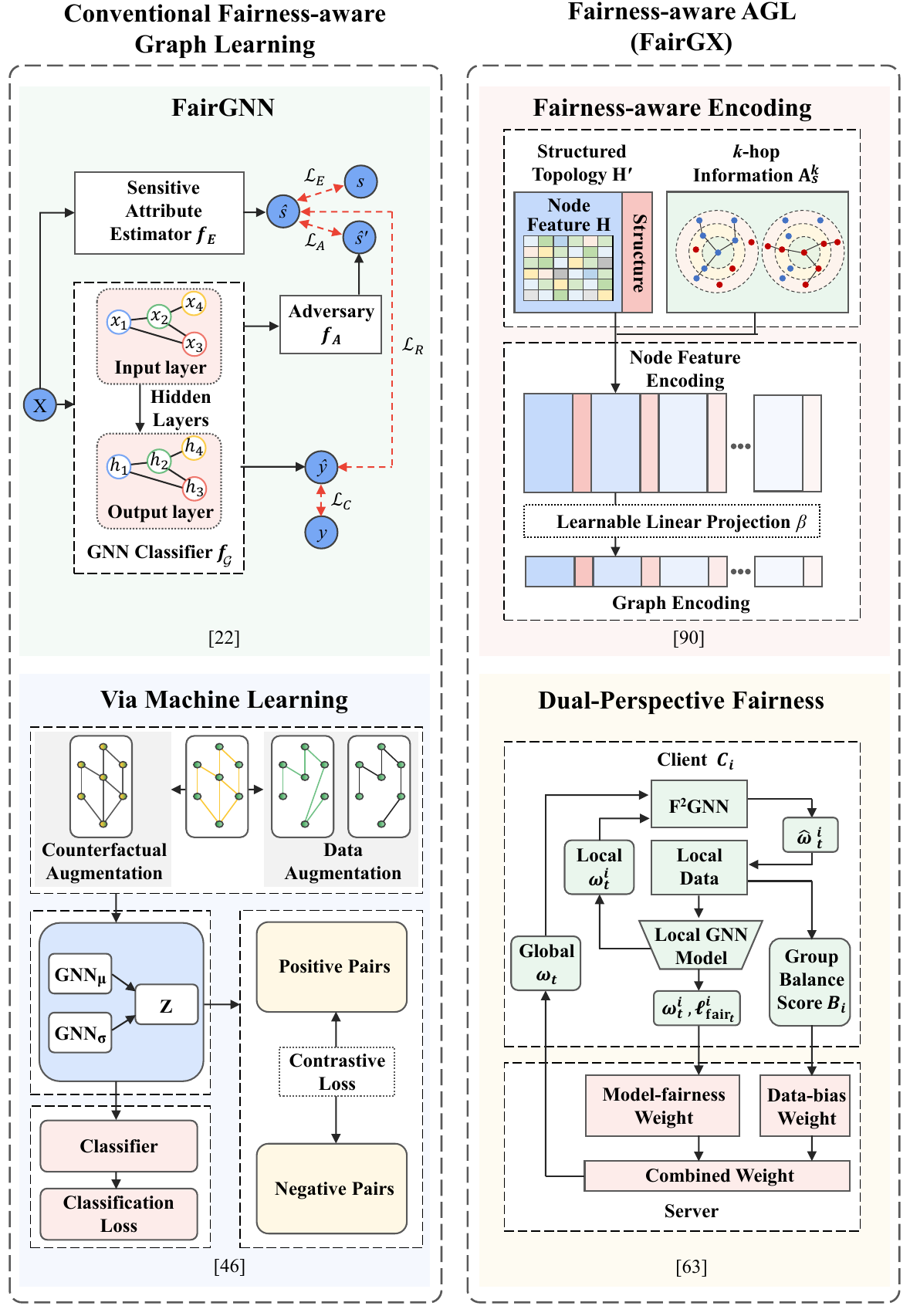}
    \caption{A comprehensive taxonomy of fairness-enhancing techniques in graph learning. 
    The taxonomy bifurcates into two primary paradigms. 
    The left side delineates conventional fairness-aware graph learning, where bias mitigation relies on specialized controllers or modular ML interventions. 
    Conversely, the right side showcases two representative categories of Fairness-aware AGL, termed FairGX in this survey. 
    Given the integration of augmentation components, these FairGX algorithms necessitate tailored debiasing strategies to counteract the specific biases emerging from the augmentation process.}
    \label{fig:MV1}
    \vspace{-2em}
\end{figure}

\par While AGL has achieved significant progress, the unique fairness challenges introduced by the augmenting ML remain largely overlooked. 
Consequently, when applied in human-centred applications, these methods may produce discriminatory outcomes for certain groups~\cite{luo2025fairness}. 
For example, in social-network-based job recommendations, individuals of a specific gender may receive fewer opportunities, or underrepresented racial groups may encounter biased results~\cite{chen2023bias}. 
As AGL adoption expands, such discrimination could permeate high-risk domains, including disaster response~\cite{xia2023cengcn}, criminal justice~\cite{feng2023criminal}, and loan approval~\cite{cheng2023regulating}, where algorithmic decisions carry profound, life-altering consequences. 
Therefore, ensuring fairness in AGL is an essential and urgent challenge.

\par While fairness in graph learning has garnered significant attention, existing surveys primarily focus on conventional settings, where fairness is pursued by modifying or constraining components within the standard graph learning~\cite{kang2022algorithmic}. 
For example, Dong et al.~\cite{dong2023fairness} explore various fairness challenges in traditional graph mining, emphasising how regulating message-passing mechanisms can enhance fairness. 
Their work also highlights the importance of mitigating biases in data representation and algorithmic decisions. 
Similarly, Dai et al.~\cite{dai2024comprehensive} analyze fairness issues in Graph Neural Networks (GNNs), underscoring the necessity of integrating fairness constraints to ensure equitable outcomes.
However, these works largely overlook the unique fairness challenges emerging from AGL, where traditional graph learning is augmented with specialized ML techniques. 
Consequently, current surveys fail to taxonomize the fairness issues and corresponding solutions in AGL settings. 
This omission presents a critical gap that impedes the development of fairness-aware methodologies, especially in scenarios where downstream tasks impose additional complexity or domain-specific requirements.

\par Compared to conventional graph learning, where fairness efforts primarily focus on regulating the message-passing process, AGL introduces two distinct challenges. 
First, fairness in AGL must encompass both the core graph learning mechanisms and the additional processes introduced by the integrated ML paradigms. 
For example, in graph federated learning, computations occur at both the local client and the server sides, involving weight aggregation and distribution. 
This dual structure creates fairness requirements on both sides that cannot be addressed by controllers designed for conventional graph learning. 
The second challenge arises from the heightened complexity of fairness issues. 
Beyond data and structural bias, AGL introduces new biases inherent to specialised ML techniques. 
For instance, the global attention mechanism in graph transformers may induce attention biases, raising further fairness concerns. 
Identifying and addressing these novel bias sources is critical for advancing AGL fairness research and ensuring equitable outcomes.

\par
\begin{figure*}[t] 
    \centering
	\includegraphics[width=0.9\textwidth]{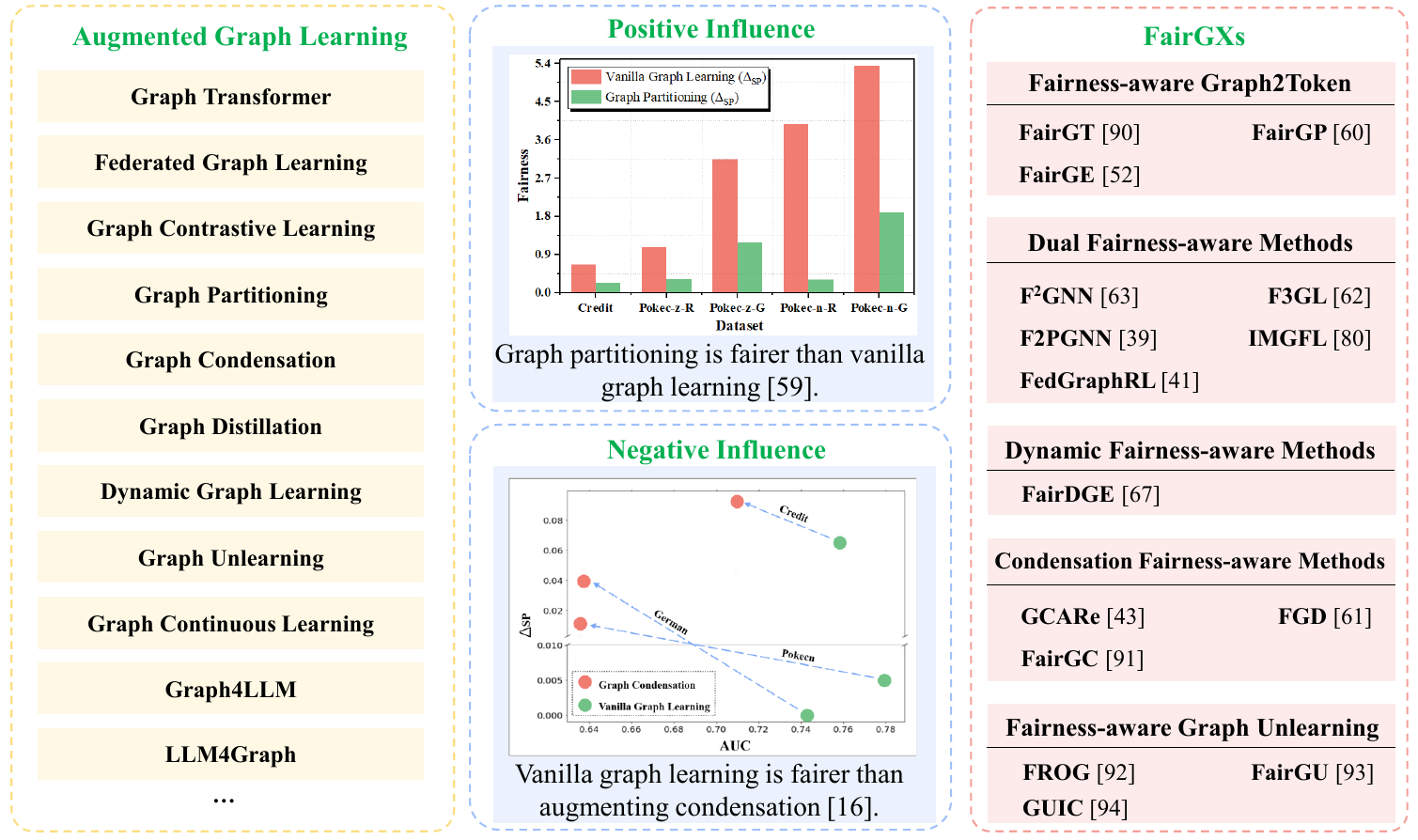}
    \caption{Overview of the AGL framework. 
    This illustration details the diverse ML augmentations integrated into graph learning, their distinct impacts on algorithmic fairness, and a systematic classification of existing literature organized by their respective augmentation techniques.}
    \label{fig:MV}
\end{figure*}

 Notably, several conventional graph learning approaches incorporate ML techniques to enhance fairness, forming a substantial body of research. 
While effective within standard framework, these methods do not adequately address the unique challenges inherent to AGL. 
To provide a comprehensive perspective, this survey first examines methods leveraging ML techniques to improve fairness in conventional graph learning. 
Subsequently, we shift our focus to AGL-specific approaches, emphasising how ML integration influences fairness and evaluating strategies developed to mitigate these challenges.
As illustrated in Figure~\ref{fig:MV1}, we establish a taxonomy that categorizes existing literature into two main groups: conventional graph learning and AGL. 
Our primary contribution lies in the systematic analysis of AGL across various scenarios, as depicted in Figure~\ref{fig:MV}.
The main contributions of this survey are summarized as follows:

\begin{itemize} [leftmargin=0.5cm]
    \item \textbf{Novel Perspective on Fairness in Graph Learning:}  
    This work provides a formal conceptualization and investigation of fairness in AGL, extending the research scope beyond conventional graph learning. 
    Unlike prior methods that focus on internal graph learning processes, we specifically examine the unique fairness challenges emerging from the integration of specialized ML techniques.
    \item \textbf{Impact Analysis of ML Augmentation:}  
    We analyze the fairness implications of incorporating diverse ML paradigms into graph-based models. 
    By identifying novel bias sources and systemic challenges inherent to AGL, this survey offers critical insights and actionable guidance for developing robust mitigation strategies in future research.
    \item \textbf{Comprehensive Taxonomy and Technical Summary:}  
    We establish a structured taxonomy of fairness-aware AGL methods, which we collectively term FairGX. 
    This overview encompasses major technical paradigms, consolidates existing research efforts, and serves as a foundational reference for advancing fairness-aware graph learning in increasingly complex ML environments.
\end{itemize}

\section{Traditional Fairness-aware Graph Learning}
\par Graph learning has demonstrated remarkable capabilities across a wide range of applications. 
However, models in this domain are prone to inheriting and amplifying biases embedded in the input data, leading to unfair or discriminatory predictions~\cite{purificato2025gnn}.
In these contexts, specific attributes (e.g., gender, race, age, and region) are legally protected to prevent abuse and are classified as sensitive attributes~\cite{dai2023learning}. 
Despite their importance, GNNs may generate biased predictions that discriminate against subgroups defined by these attributes~\cite{xu2025fairness}. 
For instance, GNNs have been shown to exhibit racial discrimination in medical underdiagnoses or gender discrimination in loan approval processes. 
Consequently, mitigating such discrimination to achieve algorithmic fairness remains a pivotal challenge in graph learning.

\par Recent advances have begun to address the limitations of conventional fairness methods by integrating sophisticated ML techniques into graph-based models to more effectively promote equity~\cite{li2025fairness}. 
These modern approaches aim not only to mitigate structural and attribute-level biases but also to preserve, or even enhance, model performance~\cite{ye2026fairgse}.
Compared to conventional constraint-based methods, they provide more comprehensive and resilient solutions by explicitly targeting the underlying sources of bias across both the data and structural dimensions. 
Prominent strategies include representation learning, adversarial debiasing, and causal inference frameworks, which enable these methods to capture intricate fairness dynamics inherent in complex graph learning environments.

\begin{table}[t]
    \centering
    \caption{Notations.}
    \begin{tabular}{p{1.2cm}p{6cm}}
    \toprule
    \textbf{Notations}          &\textbf{Definitions} \\
    \midrule
    $\mathcal{G}$ & Undirected graph \\
    $\mathcal{V}$ & Node set \\
    $\mathcal{N}_i$ & Neighbor set of node $v_i$ \\
    $\mathbf{H}$ & Node attribute matrix \\
    $\mathbf{A}$ & Adjacency matrix \\
    $\hat{\mathbf{A}}$ & Attention matrix \\
    $y$ & Ground truth label \\
    $\hat{y}$ & Predicted label \\
    $s$ & Sensitive attribute \\
    $v_i$ & Node of the graph \\
    $\mathbf{H}[i,s]$ & Sensitive attribute of node $v_i$ \\
    \bottomrule
    \end{tabular}
    \label{tab:notations}
\end{table}

\subsection{Notations}
\par Unless otherwise specified, the following mathematical conventions are consistently applied throughout this paper to ensure clarity.
We represent sets using calligraphic uppercase letters (e.g., $\mathcal{A}$).
Matrices are denoted by bold uppercase letters (e.g., $\mathbf{A}$), while vectors are indicated by bold lowercase letters (e.g., $\mathbf{x}$).
The cardinality of a set $\mathcal{A}$ is expressed as $|\mathcal{A}|$.
Adopting NumPy-style indexing, $\mathbf{A}[i,j]$ refers to the $(i,j)$-th element of matrix $\mathbf{A}$. 
Correspondingly, $\mathbf{A}[i,:]$ and $\mathbf{A}[:,j]$ characterize the $i$-th row and $j$-th column of $\mathbf{A}$, respectively.

\par A graph is formally defined as $\mathcal{G} = \{\mathcal{V}, \mathbf{A}, \mathbf{H}\}$, where $\mathcal{V}$ is the set of $n$ nodes ($|\mathcal{V}| = n$).
The adjacency matrix $\mathbf{A} \in \mathbb{R}^{n \times n}$ represents the structural connectivity, and the node attribute matrix $\mathbf{H} \in \mathbb{R}^{n \times d}$ contains $d$-dimensional features for each vertex.
For any node $v_i \in \mathcal{V}$, its set of neighbors is denoted by $\mathcal{N}_i$.
The entries of the adjacency matrix are defined such that $\mathbf{A}[i,j] = 1$ if $v_j \in \mathcal{N}_i$, and $0$ otherwise.
The vector $\mathbf{H}[:,s]$ signifies a sensitive attribute across the entire node set, whereas $\mathbf{H}[i,s]$ specifies the sensitive value of node $v_i$.
In predictive modeling, $y$ and $\hat{y}$ denote the ground-truth and predicted labels, respectively.
A comprehensive summary of these notations is provided in Table~\ref{tab:notations}.

\subsection{Preliminaries}
\par In real-world scenarios, such as online social platforms, users are often reluctant to disclose sensitive personal information (e.g., age, gender, or occupation) typically used to define protected subgroups~\cite{zhao2025fs}. 
This reluctance presents a significant challenge for fair learning, particularly in graph-based systems. 
Within this context, various fairness notions have been proposed to mitigate disparate outcomes and ensure equitable treatment.

\par In ML, fairness is typically defined along two complementary dimensions: \textbf{group fairness} and \textbf{individual fairness}, both focusing on the role of sensitive attributes or subgroup memberships~\cite{xie2022fairrankvis}.
\textbf{Group fairness} aims to eliminate systematic disparities across demographic subgroups by ensuring parity in model outputs, such as prediction rates, accuracy, or error distributions, across groups defined by sensitive attributes~\cite{chan2024group}. 
This notion is widely adopted due to its interpretability and alignment with anti-discrimination regulations.
In contrast, \textbf{individual fairness} emphasizes treating similar individuals consistently, regardless of their group membership~\cite{dong2021individual}. 
Rather than relying on predefined group labels, individual fairness assumes a similarity metric exists and requires that individuals with comparable attributes receive similar predictions~\cite{he2024on}. 
This personalized approach is particularly relevant where group boundaries are ambiguous or overlapping~\cite{zhu2025sagif}.

\par In graph learning, however, fairness considerations extend beyond sensitive attributes to encompass structural characteristics. 
This introduces \textbf{structural fairness}, which addresses biases embedded within the graph topology.
For example, in node classification, high-degree nodes (e.g., influential users or highly-cited papers) typically possess more informative neighborhoods, leading to disproportionately higher predictive performance. 
Conversely, low-degree nodes often suffer from limited contextual information, resulting in inferior utility.
Structural fairness seeks to mitigate such disparities by promoting equitable outcomes, such as embedding quality or representation utility, across nodes regardless of their degree or connectivity. 

\paragraph{Group Fairness Metrics} 
Several metrics have been adopted to quantify group-level bias.
\textbf{Statistical Parity} (SP)~\cite{dwork2012fairness}, also known as Demographic Parity or Independence, requires predictions to be independent of sensitive attributes $\mathbf{H}[:,s]$. 
The discrepancy $\Delta_\text{SP}$ measures the acceptance rate difference between two sensitive subgroups. 
For binary labels and features, it is defined as:
\begin{equation}
    \Delta_\text{SP}=|\mathbb{P}(\hat{y}=1|\mathbf{H}[i,s]=0)-\mathbb{P}(\hat{y}=1|\mathbf{H}[i,s]=1)|.
    \label{equ:delta_SP}
\end{equation}

For multi-class sensitive attributes, it can be extended using variance~\cite{jiang2023fairness}:
\begin{equation}
    \Delta_\text{SP}=\text{Var}_{j=1}^m\big(|\mathbb{P}(\hat{y}=1|\mathbf{H}[i,s]=i)\big).    
\end{equation}

\textbf{Equal Opportunity}~\cite{hardt2016equality} requires that the probability of a positive outcome for instances in the positive class be equitable across subgroups. 
Mathematically:
\begin{equation}
    \begin{aligned}
        \Delta_\text{EO} &=|\mathbb{P}(\hat{y}=1|y=1,\mathbf{H}[i,s]=0)\\&\quad-\mathbb{P}(\hat{y}=1|y=1,\mathbf{H}[i,s]=1)|.
    \end{aligned}
    \label{equ:delta_EO}
\end{equation}

\textbf{Equal Odds}~\cite{hardt2016equality} strengthens EO by requiring predictions to be independent of sensitive attributes conditional on the true label for both positive and negative classes.

\par To measure group-level fairness, especially in federated recommendation systems, a relaxation of the equalized odds condition is often adopted. 
Inspired by this principle, the following empirical loss function captures the disparity in performance across sensitive groups:
\begin{equation}
    \begin{aligned}
    &\mathcal{L}_{fair}(\textbf{M}) \\&\quad= \left| \frac{1}{|\mathbf{H}[u,s]=0|} \sum_{\mathbf{H}[u,s]=0} \textbf{M}(u) \right.\\ &\quad\quad - \left.\frac{1}{|\mathbf{H}[u,s]=1|} \sum_{\mathbf{H}[u,s]=1} \textbf{M}(u) \right|^\alpha,
    \end{aligned}
    \label{equ:fl}    
\end{equation}
where $\alpha \in \{1, 2\}$ controls the smoothness of the fairness penalty. Here, $\textbf{M}(u)$ reflects the individual performance of user $u$, and a smaller value of $\mathcal{L}_{fair}$ indicates higher group fairness. 
Minimizing $\mathcal{L}_{fair}$ while preserving overall model utility becomes the central objective in fair federated learning (FL) based recommendation systems. 
While Equation~\ref{equ:fl} considers a binary sensitive attribute, its extension to multi-group scenarios is conceptually straightforward~\cite{liu2022fairness}.

\paragraph{Individual fairness metrics} \textbf{Consistency}~\cite{lahoti2019operationalizing} quantifies how similarly a model treats similar individuals based on a similarity matrix $\mathbf{M}$:
\begin{equation}
    \operatorname{Con}_{fair}=1-\frac{\sum_i \sum_j\left|\hat{y}_i-\hat{y}_j\right| \cdot \mathbf{M}_{i j}}{\sum_i \sum_j \mathbf{M}_{i j}} \quad(i \neq j).
\end{equation}

\textbf{Fairness Homophily}~\cite{li2025fairness} captures the tendency of nodes with similar sensitive attributes to be connected. 
The ratio for node $v_i \in \mathcal{V}$ is defined as:
\begin{equation}
    h_i=\frac{|\{(v_i,v_j)|v_j \in \mathcal{N}_i \ \text{and}\  \mathbf{H}[i,s]=\mathbf{H}[j,s]\}|}{|\mathcal{N}_i|}.
\end{equation}

\par The overall fairness homophily for the graph is then obtained by averaging over all nodes:
\begin{equation}
    h_\mathcal{G} = \sum_{i=1}^n \frac{h_i}{n}.
\end{equation}

It is important to emphasize that, unlike classical homophily~\cite{chen2024polygcl} which is usually computed based on node labels (e.g., $y_i$), fairness homophily is defined with respect to sensitive attributes.
This shift in focus allows for an assessment of whether nodes with similar protected characteristics are more likely to interact, providing an additional perspective on fairness in graph topology.

\paragraph{Structural fairness metrics} \textbf{Degree Bias}~\cite{kang2022rawlsgcn} is measured by the variance of average cross-entropy loss across nodes grouped by degree $k$:
\begin{equation}
    \begin{aligned}
        \operatorname{AvgCE}(k) &= \mathbb{E}[\{\mathrm{CE}(v_i) \mid \forall v_i \text{ such that } |\mathcal{N}_i| = k\}], \\
        \operatorname{Bias} &= \operatorname{Var}(\{\operatorname{AvgCE}(k) \mid \forall k\}),
    \end{aligned}
\end{equation}
where $|\mathcal{N}_i|$ refers to the degree of node $v$ and $\mathrm{CE}(v_i)$ is the cross-entropy loss for node $v_i$.

\par Some studies use accuracy gap between different subgroups to measure the fairness, also named as degree bias.
To assess allocation fairness, Pan et al.~\cite{pan2024towards} examine two distinct dimensions: model gradient and payoff. 
Specifically, the fairness of the model gradient is quantified by the Pearson correlation coefficient $\rho(\xi,\psi)$ between self-trained test accuracy $\xi$, and collaborative accuracy $\psi$ within the FL framework~\cite{xu2021gradient}. 
Regarding payoff fairness, the allocation mechanism is evaluated by categorizing agents based on data quality (low, medium, and high). 
An effective allocation approach is characterized by a positive correlation where agents contributing superior data receive higher payoffs.

\par In the context of Federated Graph Neural Networks (FedGNNs), Xia et al.~\cite{xia2024enhancing} adopt the standard deviation of client test accuracies to measure performance equity, while utilizing average accuracy to represent overall system utility, following~\cite{wang2021federated}. 
Furthermore, Mao et al.~\cite{mao2023gcare} investigate degree bias using two complementary accuracy metrics. 
The first, $\Delta_\text{acc}$, captures the discrepancy between the highest and lowest performing subgroups. 
The second, $\delta_\text{acc}$, reflects the standard deviation of accuracy across all defined subgroups. 
Formally, given the accuracy $a_i$ for each group $i \in \mathcal{G}$, these metrics are defined as:
\begin{equation}
    \Delta_\text{acc} := \max_{i \in \mathcal{G}} a_i - \min_{j \in \mathcal{G}} a_j,
\end{equation}
\begin{equation}
    \delta_\text{acc} := \sqrt{\mathbb{E}[(a_i - \bar{a})^2]}.
\end{equation}

\par For recommendation tasks, fairness evaluation often involves ranking discrepancy metrics. 
Notably, Yang et al.~\cite{yang2017measuring} employ the average Hit Ratio Difference (rHR) and Normalized Discounted Difference (rND) across various thresholds $k \in \mathcal{K}$. 
The rHR metric measures the deviation between the Tail-to-Head (T$2$H) item ratio in the top-$k$ candidates and its global distribution. 
A smaller rHR value signifies enhanced fairness. Mathematically, rHR is expressed as:
\begin{equation}
    \operatorname{rHR} = \frac{1}{Z} \sum_{k \in \mathcal{K}} \left| \frac{|T2H_k|}{k} - \frac{|T2H|}{|\mathcal{V}|} \right|,
\end{equation}
where $|T2H_k|$ is the count of T$2$H nodes in the top-$k$ results, $|\mathcal{V}|$ is the total vertex count, and $Z$ serves as a normalization constant. 
Similarly, rND incorporates a logarithmic decay factor to penalize discrepancies in higher-ranking positions, where lower values indicate superior fairness performance:
\begin{equation}
    \operatorname{rND} = \frac{1}{Z} \sum_{k \in \mathcal{K}} \frac{1}{\log_2 k} \left| \frac{|T2H_k|}{k} - \frac{|T2H|}{|\mathcal{V}|} \right|.
\end{equation}

\subsection{Enhancing Fairness via ML}
\par Recent research has demonstrated that contrastive learning can significantly enhance fairness in graph learning by addressing the causal impact of data biases on node representations. 
By maximising consistency across diverse graph structures or counterfactual augmentations, contrastive learning ensures that nodes are treated more equitably, thereby reducing structural biases. 
This paradigm enables models to learn fairer embeddings by disentangling sensitive from non-sensitive attributes, which mitigates the influence of protected information and fosters unbiased predictions.

\par For instance, the Fair Contrastive Learning based on Counterfactual Graph Augmentation (FCLCA) framework~\cite{li2024contrastive} employs structural augmentation to generate dual counterfactual graphs. 
By maximizing representation consistency across augmented views, FCLCA  ensures equitable treatment of nodes and reduces structural bias. 
Furthermore, FCLCA incorporates adversarial debiasing to restrict the leakage of sensitive attributes into node embeddings.

\par Similarly, Fair Disentangled Graph Neural Networks (FDGNN)~\cite{zhang2025disentangled} leverage contrastive learning to isolate sensitive attributes from non-sensitive ones. 
To mitigate bias, FDGNN generates instances with identical sensitive values but distinct adjacency matrices, utilizing counterfactual augmentation to perturb sensitive attributes while preserving the graph topology.
By isolating these sensitive attributes, FDGNN directs the model's focus toward task-relevant, non-sensitive attributes, promoting equitable predictions.

\par Recent advances in graph learning have focused on improving fairness without relying on explicit demographic labels, which are often restricted by privacy regulations.
FairGKD~\cite{zhu2024the} introduces a knowledge distillation framework that utilizes multiple ``fairness experts" to guide a student model. 
This approach balances utility and fairness by distilling synthetic, fairness-aware knowledge without direct access to sensitive data.
Alternatively, FairSAD~\cite{zhu2024fair} enhances fairness through Sensitive Attribute Disentanglement, which separates sensitive information into independent components rather than eliminating it entirely. 
By employing a channel masking mechanism to decorrelate these components, FairSAD minimizes bias while preserving critical task-related information to maintain high model utility.

\subsection{The Gap between Fairness-aware Graph and Machine Learning}
\par Fairness-aware ML methods, such as adversarial debiasing, group fairness constraints, and counterfactual fairness, have made significant strides in promoting equitable outcomes in traditional settings~\cite{chen2023privacy}. 
These approaches typically operate on independent and identically distributed data, ensuring that models do not discriminate against demographic groups based on protected attributes like gender, race, or socio-economic status~\cite{ezzeldin2023fairfed, shi2024towards}. 
However, these methods often fail to account for the inherent relational structure in graph data, making them unsuitable for networked applications where dependencies between entities (nodes) crucially shape model outcomes.

\par In graph learning architectures, such as GNNs, nodes are interconnected through edges that encode complex structural dependencies~\cite{luo2026fairge}. 
These dependencies are frequently overlooked by traditional fairness metrics that treat each data point in isolation. 
For instance, standard metrics like SP or EO cannot capture biases emerging from graph topology, such as the prevalence of intra-group homophily over inter-group connectivity. 
In real-world networks, such structural bias significantly impacts fairness, as certain groups may be disadvantaged based on their topological position rather than their individual attributes. 
Consequently, applying traditional fairness-aware ML approaches to graphs often results in oversimplified pruning that balances group populations without considering the intrinsic graph geometry~\cite{zayed2024fairness}.

\par Moreover, many counterfactual fairness methods~\cite{xu2023disentangled} rely on causal assumptions that are difficult to translate into graph contexts~\cite{xu2022assessing,robertson2024fairpfn}. 
While counterfactual fairness seeks prediction consistency across altered sensitive attributes, applying this to graphs is non-trivial; modifying a single node's attribute can disrupt the systemic dependencies encoded within the topology. 
Without integrating graph structure, these methods remain inadequate for addressing fairness in heterogeneous, interconnected datasets.
To bridge this gap, new frameworks must be developed to harmonize the unique topological characteristics of graph data with advanced ML augmentation, as exemplified by the proposed FairGX taxonomy.

\section{Influence of Augmentation on Graph Learning}
\par AGL refers to the paradigm that integrates advanced machine learning techniques, external knowledge, or data enhancement strategies into traditional graph algorithms to boost representation capacity and satisfy diverse downstream requirements. 
However, this augmentation also introduces new factors that influence the fairness of graph learning models. 
It is therefore critical to systematically investigate these effects in order to improve fairness in AGL and to identify the underlying causes of fairness-related concerns. 
When such influences are beneficial, efforts can be directed toward amplifying their positive impact. 
Conversely, if adverse effects are observed, targeted strategies should be developed to mitigate or eliminate the resulting unfairness.

\par Distinct from traditional ML for fairness, AGL primarily focuses on integrating sophisticated ML algorithms into the graph computing pipeline to enhance performance, generalizability, and scalability.
Formally, this integration creates a deep structural-functional coupling in the augmented model $f(\mathcal{G}, \mathbf{X}; \Theta)$. 
In this context, the model transcends a simple extension of graph neural networks to become a hybrid system, where ML-driven dynamics and graph topological dependencies are inextricably intertwined.
Consequently, conventional graph fairness methods, which typically assume static or standard graph distributions, become suboptimal or even invalid as they fail to account for the new bias patterns introduced by the augmentation process. 
Simultaneously, standard ML fairness techniques remain structure-agnostic, rendering them incapable of addressing the biased information propagation inherent in graph structures.

\subsection{Positive Influence}
\par Some ML techniques have been employed to enhance fairness in graph learning. 
Researchers have observed that these techniques, when applied to address specific limitations or requirements, can also contribute to improving fairness.

\par For example, contrastive learning has reduced the dependency on human annotations, establishing Graph Contrastive Learning (GCL) as a promising self-supervised approach for node representation learning. Studies indicate that GCL tends to exhibit greater fairness compared to graph convolutional learning. 
A theoretical explanation is provided by~\cite{wang2022uncovering}, which attributes this fairness to the intra-community concentration and inter-community scatter properties of GCL.
These properties create clearer community structures, helping low-degree nodes remain distant from community boundaries. 
Furthermore, their experiments reveal that GCL methods outperform Graph Convolutional Networks (GCNs) in maintaining degree bias, evidenced by a smaller performance gap between tail nodes and head nodes. 
This finding highlights the potential of GCL to enhance degree bias in graph learning.

\par In graph learning, particularly in graph transformers, higher-order nodes (i.e., nodes with high degrees) often dominate the learning process (named over-globalizing)~\cite{xing2024less, shehzad2024graph}, which can amplify the influence of sensitive attributes associated with these nodes, leading to bias in the model.
This dominance can lead to the amplification of sensitive attribute correlations present in these nodes, ultimately inducing bias in the model's predictions.
As shown in the work of Luo et al.~\cite{luo2025fairgp}, this phenomenon has significant fairness implications. Specifically, the model tends to overfit to the structural and attribute patterns of high-degree nodes, leading to skewed outcomes that favor the majority group represented among them. 
This issue is particularly evident in augmented graph learning models such as graph transformers, where attention mechanisms and message passing are heavily influenced by node connectivity patterns, further amplifying the impact of high-degree nodes.

\par Empirical validation of this bias was conducted by analyzing the distribution of sensitive attributes across different benchmark datasets, as shown in Table~\ref{tab:pos_inf}. 
In these experiments, sensitive attributes were defined as binary attributes (e.g., gender or race), and their distribution was compared between the overall node set and the subset of higher-order nodes. 
The results were visualized using color-coded bars: \textcolor{red}{red} indicating cases where the majority class had sensitive attribute = 1 ($|\textbf{H}[i,s] = 1| > |\textbf{H}[i,s] = 0|$), and \textcolor{blue}{blue} where it had sensitive attribute = 0 ($|\textbf{H}[i,s] = 1| < |\textbf{H}[i,s] = 0|$).
Specifically, if $|\textbf{H}[i,s] = 1| = |\textbf{H}[i,s] = 0|$, it will show the black "1".
These visualizations revealed a clear pattern, when the sensitive attribute distribution among higher-order nodes deviated from that of the full graph, graph transformers tended to learn and make predictions aligned with the higher-order node distribution. 
In some cases, this misalignment led to predictions that were directly opposite to what would be expected based on the overall graph distribution.

\begin{table}[t]
    \centering
    \tabcolsep=0.15cm
    \caption{Distribution of different sensitive attributes. \textbf{\textcolor{red}{Red}} indicates $|\textbf{H}[i,s] = 1| > |\textbf{H}[i,s] = 0|$, and \textbf{\textcolor{blue}{blue}} indicates cases $|\textbf{H}[i,s] = 1| < |\textbf{H}[i,s] = 0|$.}
    \begin{tabular}{lccc}
        \toprule
        \textbf{Datasets} & \textbf{All Nodes} & \textbf{Higher-order} & \textbf{Prediction ($\hat{y}=1$)} \\
        \midrule
        \textbf{Credit} & \textbf{\textcolor{red}{10.17}} & \textbf{\textcolor{red}{2.25}} & \textbf{\textcolor{red}{2.94}} \\
        \textbf{Pokec-z-R} & \textbf{\textcolor{red}{1.82}} & \textbf{\textcolor{blue}{1.29}} & \textbf{\textcolor{blue}{2.63}} \\
        \textbf{Pokec-z-G} & 1 & \textbf{\textcolor{blue}{1.25}} & \textbf{\textcolor{blue}{3.10}} \\
        \textbf{Pokec-n-R} & \textbf{\textcolor{red}{2.12}} & \textbf{\textcolor{red}{2.51}} & \textbf{\textcolor{red}{3.50}} \\
        \textbf{Pokec-n-G} & \textbf{\textcolor{red}{1.26}} & \textbf{\textcolor{blue}{1.37}} & \textbf{\textcolor{blue}{2.69}} \\
        \textbf{AMine-L} & \textbf{\textcolor{red}{1.18}} & \textbf{\textcolor{blue}{1.36}} & \textbf{\textcolor{blue}{6.81}} \\
        \bottomrule
    \end{tabular}
    \label{tab:pos_inf}
\end{table}

\par Even when the majority group in both distributions (overall and higher-order nodes) was the same, the model's predictions more closely reflected the proportion of sensitive attributes among the higher-order nodes rather than the graph as a whole.
This observation indicates that the demographic composition of high-degree nodes significantly affects the fairness of graph transformer-based models.

\par Graph partitioning mitigates this issue by limiting the influence of high-degree nodes through localized learning within smaller, more balanced subgraphs.
As illustrated in Fig.~\ref{fig:MV}, partitioning the graph into subsets with more uniform degree distributions and sensitive attribute compositions reduces the dominance of structurally influential nodes, thereby promoting fairer learning outcomes.
Thus, in the context of AGL models such as graph transformers, graph partitioning serves not only as a tool for improving scalability but also as an effective mechanism for enhancing fairness.

\par Thus, the augmentation of ML techniques such as contrastive learning and graph partitioning in graph learning methods has been shown to yield fairer outcomes, demonstrating their potential to address fairness challenges while meeting other objectives.

\subsection{Negative Influence}
\par Some ML techniques have fulfilled specific downstream task requirements but have also modified the structure of graph learning, which has adverse effects on fairness.

\par For instance, graph condensation, which is designed to reduce the computational complexity of graph learning and improve model interpretability, has been shown to introduce substantial fairness concerns~\cite{mao2023gcare}. 
Although state-of-the-art condensation techniques effectively distill representative information from original graphs, they tend to perform unevenly across different demographic subgroups, leading to amplified disparities in model outcomes~\cite{feng2023fair}.

\par In particular, when GNNs are trained on condensed graphs instead of full graphs, the performance gap between advantaged and disadvantaged groups often widens. 
This is especially problematic in high-stakes applications where equitable performance is crucial. 
Methods such as GCond and DosCond, while achieving comparable overall classification accuracy to models trained on original graphs, consistently exhibit a bias in favor of advantaged groups. 
This results in a significant drop in predictive performance for disadvantaged groups, exacerbating existing inequalities.

\par Table~\ref{tab:neg_inf} provides a detailed comparison of group fairness, measured by $\Delta_\text{SP}$, between models trained on real graphs and those trained on condensed graphs across various graph learning and debiasing methods. 
Notably, across all evaluated settings, the distilled graphs consistently lead to lower fairness scores. 
This suggests that condensation may preserve structural or statistical patterns that are more representative of majority or high-degree nodes, while underrepresenting minority or under-connected nodes.

\begin{table}[t]
    \centering
    \tabcolsep=0.15cm
    \caption{Comparative analysis of group fairness across original and distilled graphs: An evaluation of diverse graph learning architectures and debiasing frameworks. 
    Entries highlighted in \textbf{\textcolor{red}{Red}} denote a regression in fairness performance.}
    \begin{tabular}{lcccc}
        \toprule
        \textbf{Graph} & \textbf{Real} & \textbf{Condensed} & \textbf{Real} & \textbf{Condensed} \\
        \textbf{Metrics} & $\Delta_\text{SP}$(\%) $\downarrow$ & $\Delta_\text{SP}$(\%) $\downarrow$ & $\Delta_\text{EO}$(\%) $\downarrow$ & $\Delta_\text{EO}$(\%) $\downarrow$ \\
        \midrule
        \textbf{GCN}        & 5.41    & \textbf{\textcolor{red}{12.04}}  & 3.12    & \textbf{\textcolor{red}{9.58}} \\
        \textbf{SGC}        & 2.78    & \textbf{\textcolor{red}{9.58}}   & 1.19    & \textbf{\textcolor{red}{7.14}} \\
        \textbf{GraphSAGE}  & 9.55    & \textbf{\textcolor{red}{8.77}}   & 3.92    & \textbf{\textcolor{red}{6.72}} \\
        \textbf{EDITS}      & 2.73    & \textbf{\textcolor{red}{6.34}}   & 1.82    & \textbf{\textcolor{red}{2.84}} \\
        \textbf{FairGNN}    & 6.24    & \textbf{\textcolor{red}{12.33}}  & 3.74    & \textbf{\textcolor{red}{9.12}} \\
        \textbf{InFoRM}     & 0.43    & \textbf{\textcolor{red}{17.42}}  & 0.13    & \textbf{\textcolor{red}{13.51}} \\
        \bottomrule
    \end{tabular}
    \label{tab:neg_inf}
\end{table}

\par These findings indicate that, despite their utility in model compression and efficiency gains, current graph condensation techniques come with non-negligible fairness trade-offs. 
The underlying compression objectives, which often prioritize information retention based on frequency or influence, may inadvertently overlook the representational needs of minority subgroups. 
The learned representations tend to disproportionately benefit advantaged nodes during downstream learning, reinforcing biases and leading to systematically unfair outcomes.
Therefore, it is imperative to develop fairness-aware condensation strategies that explicitly consider subgroup representation during the distillation process. Without such considerations, the deployment of graph condensation in fairness-sensitive contexts could further marginalize underrepresented groups, undermining the broader goal of equitable ML.

\par Similarly, centralised training of GNNs is often limited by privacy concerns and regulatory restrictions, making FL an attractive alternative for distributed settings. 
However, FL introduces new fairness challenges.
Biases present in local training data are readily transferred to local GNN models and are further propagated and amplified in the global model during aggregation. 
Recent studies have shown that even under centralised training, GNNs can produce unfair predictions and exhibit discriminatory behaviour~\cite{cui2023equipping}. 
In a federated learning framework, this issue becomes more pronounced, as biases from multiple local models compound during global aggregation, resulting in significant fairness concerns.

\par These examples demonstrate that certain ML techniques used in graph learning, particularly those involving structural modifications or distributed training, can exacerbate fairness issues. 
Addressing these challenges is essential for ensuring equitable outcomes in AGL.

\subsection{Novel Fairness Metrics}
\par In FedGNNs, the decentralized nature of training poses significant challenges for fairness evaluation.
Unlike centralized settings, where sensitive demographic information is often available to the model, the privacy-preserving design of FL prohibits direct access to clients' sensitive attributes on the server side.
This limitation renders fairness-aware ML techniques, many of which rely on global knowledge, inapplicable in their original form.
Consequently, there is a pressing need for fairness metrics that are compatible with the distributed and privacy-constrained architecture of FL.

\par To address this issue, Agrawal et al.~\cite{agrawal2024no} propose a formulation of user-level group fairness tailored specifically for FL settings.
Let $\mathcal{R}_u$ denote the recommendation list generated for user $u$, and let $\mathbf{M}$ represent a performance evaluation metric, such as accuracy or ranking quality.
Assume users are divided into two disjoint sensitive groups, identified by the sensitive attribute matrix $\mathbf{H}$ where $\mathbf{H}[u, s] \in \{0, 1\}$.
Group fairness is then defined by enforcing equality of expected performance across the two groups:
\begin{equation}
\mathbb{E}_u[\textbf{M}(\mathcal{R}_u)\mid \mathbf{H}[u,s]=0] = \mathbb{E}_u[\textbf{M}(\mathcal{R}_u)\mid \mathbf{H}[u,s]=1].
\end{equation}

\par In the context of GNNs, fairness is typically evaluated by measuring the independence of model outputs from sensitive attributes, often using criteria such as statistical parity or equality of opportunity~\cite{kang2021fair}.
When applied to FedGNNs, however, fairness must be considered from both local and global perspectives.
Local fairness pertains to each individual client and examines whether the model behaves fairly within its local dataset, particularly important given the non-IID (non-identically distributed) nature of client data, which may contain skewed distributions of sensitive attributes.
Global fairness, on the other hand, assesses the fairness of the aggregated global model after communication rounds, ensuring that collective predictions across all clients do not disproportionately disadvantage any sensitive group.

\par
\begin{figure}[t]
    \centering
    \subfigure[Ensuring fairness at client-level does not necessarily guarantee fairness at the server-level.]{
        \begin{minipage}[b]{0.45\textwidth}
            \centering
            \includegraphics[width=1\textwidth]{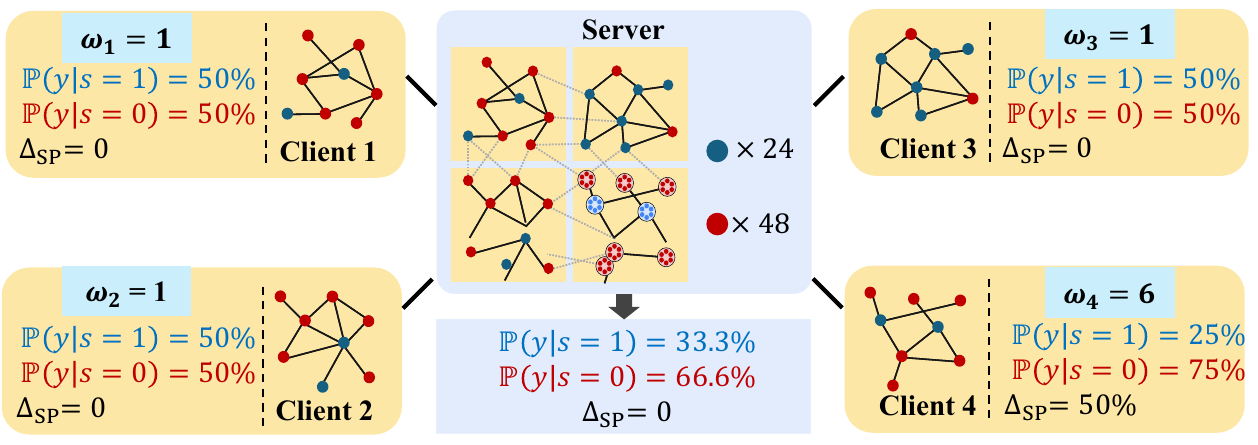}
        \end{minipage}
    }
    \subfigure[Ensuring fairness at the server-level does not necessarily guarantee fairness at the client-level.]{
        \begin{minipage}[b]{0.45\textwidth}
            \centering
            \includegraphics[width=1\textwidth]{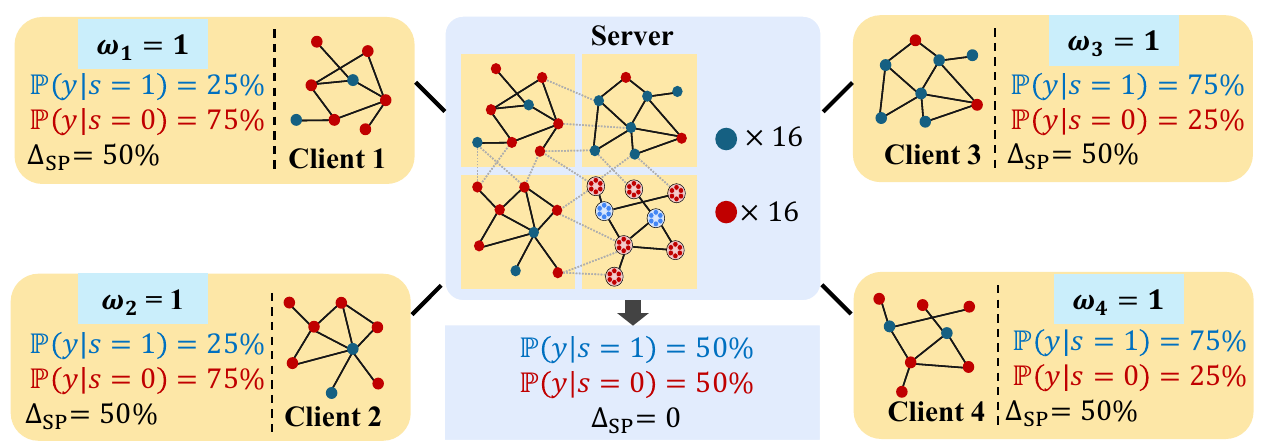}
        \end{minipage}
    }
    \caption{A toy example illustrating the disparity between local and global fairness. 
    The figure demonstrates that individual or local-level fairness interventions do not inherently translate to system-wide equity; thus, ensuring fairness from one perspective does not necessarily guarantee fairness from another.}
    \label{fig:fed}
\end{figure}

 Together, these perspectives highlight the need for fairness metrics that not only capture performance disparities but also align with the decentralized and privacy-preserving objectives of federated graph learning.
Importantly, achieving fairness at either the local or global level in isolation is not sufficient; the two perspectives are complementary but not interchangeable.
For example, even if each local model achieves fairness within its own data distribution, the aggregated global model may still exhibit unfair behavior across the broader population~\cite{luo2026utility}.
Moreover, as federated learning jointly optimizes server-side aggregation and client-side objectives, varying fairness weights ($\omega$) across different stages may create conflicting optimization goals, making it difficult for both the global model and local models to achieve fairness simultaneously.
This discrepancy arises due to the heterogeneity of client data and the complex interactions introduced during the model aggregation process.
As illustrated in Fig.~\ref{fig:fed}, such misalignments can lead to fairness degradation at the global level, despite locally fair decisions.

\par This phenomenon is further supported by theoretical analysis from Wang et al.~\cite{wang2023mitigating}, who demonstrate that global fairness cannot be represented as a linear combination of local fairness objectives.
In other words, fairness achieved independently at each client does not guarantee fairness after model aggregation, thereby exposing a structural gap in fairness transferability under the federated paradigm.

\par These insights underscore the inherent complexity of optimizing fairness in FedGNNs and emphasize the necessity of multi-level fairness evaluation.
To address this challenge, current research must develop intervention strategies that jointly consider local heterogeneity and global equity, ensuring that fairness is preserved throughout the learning pipeline, from decentralized training to global inference.

\subsection{Datasets}
\par This section summarizes adopted datasets in fairness-aware graph learning, spanning domains such as recommendation systems, social networks, citation networks, and biological and medical graphs. 
Each dataset is characterized by its graph structure, prediction task, and sensitive attributes used to evaluate fairness.
The details are shown in Table~\ref{tab:datasets}.

\begin{table*}
    \centering
    \caption{A summary of datasets used in FairGX.}
    \renewcommand{\arraystretch}{1.25}
    \begin{tabular}{llllll}
        \toprule
         \textbf{Dataset} & \textbf{Cite} & \textbf{Tasks} & \textbf{Augmenting Method} & \textbf{Sensitive Attribute} & \textbf{Fairness Types}\\
         \midrule
         MovieLens-100K & \cite{harper2015the} & Recommender System & Federated Learning & Gender & Group Fairness \\
         MovieLens-1M & \cite{harper2015the} & Recommender System & Federated Learning & Gender & Group Fairness \\
         Amazon-Movies & \cite{ni2019justifying} & Recommender System & Federated Learning & None & Degree Bias \\
         MovieLens-10M & \cite{harper2015the} & Recommender System & Dynamic Learning & Gender & Group Fairness \\
         Amazon-Books & \cite{li2024toward} & Recommender System & Dynamic Learning & None & Degree Bias \\
         GoodReads & \cite{li2024toward} & Recommender System & Dynamic Learning & None & Degree Bias \\
         Pokec-z-G & \cite{takac2012data} & Node Classification & Transformer & Gender & Group Fairness \\
         Pokec-z-R & \cite{takac2012data} & Node Classification & Transformer/Federated Learning/Condensation & Region & Group Fairness \\
         Pokec-n-G & \cite{takac2012data} & Node Classification & Transformer & Gender & Group Fairness \\
         Pokec-n-R & \cite{takac2012data} & Node Classification & Transformer/Federated Learning/Condensation & Region & Group Fairness \\
         AMiner-L & \cite{wan2019aminer} & Node Classification & Transformer & Affiliation & Group Fairness \\
         NBA & \cite{dai2023learning} & Node Classification & Transformer & Nationality & Group Fairness \\
         German & \cite{asuncion2007uci} & Node Classification & Transformer/Condensation & Gender & Group Fairness \\
         Bail/Recidivism & \cite{jordan2015effect} & Node Classification & Transformer/Condensation & Race & Group Fairness \\
         Credit & \cite{yeh2009the} & Node Classification & Transformer/Condensation & Age & Group Fairness \\
         Income & \cite{asuncion2007uci} & Node Classification & Transformer & Race & Group Fairness \\
         Cora & \cite{kipf2017semi} & Node Classification & Condensation/Unlearning & None & Degree Bias \\
         Ogbn-arxiv & \cite{hu2020open} & Node Classification & Condensation & None & Degree Bias  \\
         Citeseer & \cite{bojchevski2018deep} & Node Classification & Unlearning & None & Degree Bias \\
         Pubmed & \cite{bojchevski2018deep} & Node Classification & Unlearning & None & Degree Bias \\
         Facebook & \cite{leskovec2012learning} & Node Classification & Unlearning & Gender & Group Fairness \\
         PROTEINS &\cite{borgwardt2005protein} & Graph Classification & Federated Learning & None & Degree Bias \\
         DD &\cite{dobson2003distinguishing} & Graph Classification & Federated Learning & None & Degree Bias \\
         IMDB-BINARY &\cite{yanardag2015deep} & Graph Classification & Federated Learning & None & Degree Bias \\
         HAM10k &\cite{tschandl2018the} & Graph Classification & Federated Learning & None & Degree Bias \\
         Fed-DRG &\cite{xia2024enhancing} & Graph Classification & Federated Learning & None & Degree Bias \\
         \bottomrule
    \end{tabular}
    \label{tab:datasets}
\end{table*}

\textbf{MovieLens}~\cite{harper2015the} contains three versions, ML-100K, ML-1M, and ML-10M, each varying in scale. 
In these datasets, users and movies form a bipartite graph, where edges represent user-item interactions (ratings), and edge weights indicate rating scores. 
ML-100K consists of $943$ users, $1,682$ items, and $100,000$ ratings with values ranging from $1$ to $5$. 
ML-1M includes $6,040$ users, $3,706$ items, and $1,000,209$ ratings. 
ML-10M contains $2,342$ users, $5,579$ items, and $477,419$ edges, and is often used in dynamic graph learning tasks with timestamps from Jan $7$, $2001$ to Jan $7$, $2003$.
Fairness-aware methods in this context primarily focus on addressing gender discrimination within the recommender systems.

\textbf{Amazon-Movies}~\cite{ni2019justifying} comprises $5,515$ users, $13,509$ items, and $484,141$ ratings, with score levels ranging from $1$ to $5$.
Similar to MovieLens, users and movies form a bipartite graph, where edges represent user-item interactions (ratings), and edge weights correspond to the rating scores.
However, this dataset does not include any sensitive demographic attributes.
Thus, the debiasing attribute "user activity" is derived based on an average rating threshold.

\textbf{Amazon-Books}~\cite{li2024toward} is tailored for dynamic graph learning tasks, spanning from Aug $9$, $2006$ to Aug $12$, $2007$. 
It includes $4,054$ users, $17,651$ items, and $447,419$ interactions.
In this graph, users and items serve as nodes, while their interactions form the edges.
This dataset has been anonymized and does not include sensitive personal attributes.
Fairness-aware approaches applied to this dataset primarily address degree bias by mitigating performance disparities between high-degree and low-degree node groups in downstream graph learning tasks.

\textbf{GoodReads}~\cite{li2024toward} originates from an online book community, focusing on children’s books from Jan $1$, $2013$ to Dec $30$, $2015$, it comprises $16,701$ users, $20,823$ items, and $1,084,781$ edges.
Similar to Amazon-Books, users and items are represented as nodes, and interactions constitute the edges. 
This dataset is also anonymized. 
Fairness-aware techniques here concentrate on ensuring structural equity by addressing imbalances in learning effectiveness between nodes of varying connectivity levels.

\textbf{Pokec-z} and \textbf{Pokec-n}~\cite{takac2012data} are social network datasets extracted from different regions in Slovakia. 
Users are represented as nodes and friendships as edges. 
Pokec-z includes $67,797$ nodes and $882,765$ edges; Pokec-n includes $66,569$ nodes and $729,129$ edges. 
Each has two variants: one using region ("-r") and another using gender ("-g") as the sensitive attribute. 
The classification label is the user’s working field.
Fairness-aware studies focus on debiasing sensitive attributes.

\textbf{Credit}~\cite{yeh2009the} is a financial dataset in which edges encode user similarity based on spending and payment patterns. 
It contains $30,000$ users and $137,377$ similarity-based links. 
Age serves as the sensitive attribute, while the task is to predict credit default in the following month.

\textbf{AMiner-L}~\cite{wan2019aminer} is a co-authorship network constructed from the AMiner database. 
Each node represents a researcher, and edges denote co-authorship relations. 
The dataset includes $129,726$ users and $591,039$ edges.
The sensitive attribute is the continent of institutional affiliation, and the target is to predict the primary research field.

\textbf{NBA}~\cite{dai2023learning} includes player statistics and social connections (e.g., Twitter follows) from the $2016$–$2017$ season. 
The dataset consists of $403$ players and $16,570$ social links. 
The sensitive attribute is nationality (U.S. or overseas), and the task is to predict whether a player’s salary exceeds the median.

\textbf{Bail/Recidivism}~\cite{jordan2015effect} involves defendants released on bail in U.S. state courts from $1990$ to $2009$.
Nodes represent defendants and are connected based on shared criminal history and demographic attributes. 
The task is to predict recidivism (violent or non-violent), with race as the sensitive attribute. 
The graph includes $18,876$ nodes and $321,308$ edges.

\textbf{German}~\cite{asuncion2007uci} is a credit dataset extracted from the UCI Adult Data Set. 
It represents clients of a German bank as nodes connected via credit account similarity. 
It contains $1,000$ nodes and $22,242$ similarity links.
The goal is to classify credit risk (high or low), with gender as the sensitive attribute. 

\textbf{Income}~\cite{asuncion2007uci} is also derived from the UCI Adult Data Set, where nodes represent individuals and edges reflect social similarity, following \cite{agarwal2021towards}. 
The dataset has $14,821$ nodes and $100,483$ edges.
The task is to predict whether an individual earns more than $50,000$ annually, using race as the sensitive attribute. 

\textbf{Cora}~\cite{kipf2017semi} is a citation network consisting of computer science publications. 
In this dataset, each node corresponds to a publication, and directed edges represent citation relationships between papers.
The dataset contains $2,708$ nodes and $5,429$ edges. 
In fairness-aware graph unlearning tasks, structural attributes such as node degree or paper category are often adopted as proxies for sensitive attributes. 
Consequently, research on this dataset typically emphasizes degree bias by aiming to reduce performance disparities arising from topological imbalances.

\textbf{Ogbn-arxiv}~\cite{hu2020open} is a large-scale citation network from the Microsoft Academic Graph, containing $169,343$ nodes and $1,116,243$ edges. 
As with Cora, node degree is commonly used to partition debiasing groups, serving as a surrogate for sensitive attributes. 
Therefore, studies on this dataset also primarily focus on degree bias in graph learning, particularly in mitigating the influence of graph topology on algorithmic performance.

\textbf{Citeseer}~\cite{bojchevski2018deep} is a citation network where nodes represent documents and edges correspond to citations. 
In fairness-aware graph unlearning tasks, the paper category is often considered a sensitive attribute. 
Additionally, node degree is widely adopted to define debiasing groups, serving as a structural proxy for fairness. 
Consequently, research on this dataset primarily targets degree bias, aiming to mitigate the impact of topological disparities on model performance.

\textbf{Pubmed}~\cite{bojchevski2018deep} is another citation network, similar in format and objective to CiteSeer, where nodes represent biomedical papers and edges reflect citation links. 
Here, the paper category also functions as the sensitive attribute in fairness evaluations. 
Following practices established in Cora and CiteSeer, node degree is utilized to group nodes for fairness-aware analysis. 
As a result, studies using Pubmed predominantly address degree bias concerns, focusing on reducing performance inequalities caused by graph topology.

\textbf{Facebook}~\cite{leskovec2012learning} is an ego-network dataset constructed from anonymized Facebook user data. 
It includes $1,045$ nodes and $53,498$ edges.
Nodes are users with profile-based attributes, and edges denote friendships. 
Gender is treated as the sensitive attribute. 

\textbf{PROTEINS}~\cite{borgwardt2005protein} is a biological dataset where each graph represents a protein structure, with nodes denoting amino acids and edges indicating $3$D spatial proximity within $6$ angstroms. 
The dataset comprises $1,113$ graphs, each with an average of $39.06$ nodes and $72.82$ edges. 
Fairness-aware methods on this dataset primarily address degree bias, referring to the accuracy disparity between subgroups of varying clients. 
While the impact of statistical heterogeneity in client-specific data on convergence speed is widely recognized in federated settings, its influence on fairness, manifesting as inconsistent accuracy across clients, remains frequently overlooked.

\textbf{DD}~\cite{dobson2003distinguishing} is a molecular graph dataset in which nodes represent atoms and edges correspond to chemical bonds. 
It includes $1,178$ graphs, with an average of $284.32$ nodes and $715.66$ edges per graph. 
Node attributes include atom types and partial charges. 
Fairness-aware studies target degree bias, aiming to reduce accuracy gaps between different clients. 
Although statistical heterogeneity is known to slow global model convergence, its role in inducing biased global models, resulting in performance variance across clients, is often underestimated.

\textbf{IMDB-BINARY}~\cite{yanardag2015deep} models actor collaboration networks, where each graph represents an actor’s ego-network based on co-appearances in films. 
The dataset consists of $1,000$ graphs, with an average of $19.77$ nodes and $95.53$ edges per graph. 
Fairness-aware approaches focus on mitigating client-induced performance disparities. 
As with federated scenarios, the overlooked consequence of local heterogeneity is the emergence of biased global models that unfairly favor certain client distributions.

\textbf{HAM10k}~\cite{tschandl2018the} is a skin lesion classification dataset collected from four medical institutions, commonly used in federated medical imaging research. 
Fairness-aware methods here also emphasize reducing degree bias, which reflects the uneven diagnostic performance across different client datasets. 
While heterogeneity in local data is known to affect training dynamics, its adverse effects on fairness, such as client-level accuracy variance, are often insufficiently addressed.

\textbf{Fed-DRG}~\cite{xia2024enhancing} comprises data collected from six public datasets (APTOS~\cite{maggie2019aptos}, DeepDR~\cite{liu2022deepdrid}, FGADR~\cite{zhou2020benchmark}, e-ophtha~\cite{decenciere2013teleophta}, IDRiD~\cite{porwal2018indian}, Messidor~\cite{abramoff2013automated}) for diabetic retinopathy grading.
It is designed to evaluate both fairness and privacy in federated learning scenarios. 
Similar to the previous datasets, fairness-aware research on Fed-DRG targets degree bias and performance variability across non-identically distributed clients. 
The subtle but significant fairness risks posed by such heterogeneity call for greater attention in model development and evaluation.

\section{Fairness-aware Augmented Graph Learning}
\par AGL introduces new algorithmic mechanisms that fundamentally change how graph learning operates. 
These innovations make traditional fairness methods, originally designed for graph message-passing, ineffective. 
Fairness constraints that work well with message-passing processes cannot be directly applied to these augmented methods, as the underlying mechanisms are significantly different. 
Therefore, ensuring fairness in AGL requires the development of novel fairness-aware methods that are specifically designed to accommodate the new mechanisms introduced by ML. 
These methods must address the unique challenges posed by AGL, such as bias introduced by ML components, while maintaining or improving performance. 
Without these new methods, ensuring fairness in the context of AGL would remain a major challenge.

\par To unify these emerging research efforts, we define \textbf{FairGX} as a specialized class of fairness-aware augmented graph learning frameworks.
These methods must extend beyond traditional fairness techniques by specifically addressing the unique challenges inherent in AGL. 
Such challenges include algorithmic biases arising from sophisticated model training, skewed data distributions, and the intricate interdependencies between graph topologies and ML components.
By advancing the FairGX paradigm, researchers can ensure that AGL models uphold rigorous fairness standards across diverse downstream tasks, such as classification, recommendation, and link prediction. 
This framework remains essential even as ML augmentations fundamentally reshape the structural and functional foundations of graph learning.
\begin{table*}
    \centering
    \caption{A taxonomy of FairGX across different augmentation types.}
    \renewcommand{\arraystretch}{1.25}
    \begin{tabular}{lllll}
        \toprule
         \textbf{Augmentation Type} & \textbf{Methods} & \textbf{Cite} & \textbf{Downstream Tasks} & \textbf{Links} \\
         \midrule
         \multirow{3}{*}{Transformer (Fairness-aware Graph2Token)} & FairGT & \cite{luo2024fairgt} & Node Classification & https://github.com/LuoRenqiang/FairGT \\
         & FairGP & \cite{luo2025fairgp} & Node Classification & https://github.com/LuoRenqiang/FairGP \\
         & FairGE & \cite{luo2026fairge} & Node Classification & https://github.com/LuoRenqiang/FairGE \\\midrule
         \multirow{4}{*}{Federated Learning (Dual Fairness-aware Methods)}  & F$^2$GNN & \cite{cui2023equipping} & Node Classification & https://github.com/yuening-lab/F2GNN \\
         & F3GL & \cite{luo2026utility} & Node Classification & None \\
         & F2PGNN & \cite{agrawal2024no} & Recommender System & https://github.com/nimeshagrawal/F2PGNN \\
         & IMGFL & \cite{pan2024towards} & Graph Classification & https://github.com/zjunet/FairGraphFL \\
         & FedGraphRL & \cite{xia2024enhancing} & Graph Classification & None \\
         \midrule
         Dynamic Learning (Dynamic Fairness-aware Methods) & FairDGE & \cite{li2024toward} & Recommender System & https://github.com/Abigale001/FairDGE \\\midrule
         \multirow{3}{*}{Condensation (Fairness-aware Condensation Methods)}  & GCARe & \cite{mao2023gcare} & Node Classification & None \\
         & FGD & \cite{feng2023fair} & Node Classification & None \\
         & FairGC & \cite{gao2026fairgc} & Node Classification & https://github.com/Luorenqiang/FairGC \\\midrule
         \multirow{3}{*}{Unlearning (Fairness-aware Graph Unlearning)}  & FROG & \cite{chen2025frog} & Node Classification & None \\
         & FairGU & \cite{luo2026fairgu} & Node Classification & https://github.com/Luorenqiang/FairGU \\
         & GUIC & \cite{wang2026guic} & Node Classification & None \\
         \bottomrule
    \end{tabular}
    \label{tab:taxonomy}
\end{table*}

\subsection{Fairness-aware Graph2Token}
\par Graph transformers encode graph information without relying on traditional message-passing mechanisms, rendering conventional fairness-aware graph learning methods inadequate for addressing fairness concerns in graph transformers. 
Instead, graph information must first be converted into a sequence of tokens through a graph tokenization process, commonly known as a tokenizer or Graph2Token module, before being processed by the transformer.
This additional representation learning stage may introduce new sources of bias and fairness risks that are absent in traditional graph neural networks.

\par To address this limitation, FairGT, a fairness-aware encoding method, is introduced by~\cite{luo2024fairgt}, specifically designed to mitigate fairness issues in graph transformers. 
FairGT employs a structural attribute selection strategy and a multi-hop node attribute integration method to ensure the independence of sensitive attributes and enhance fairness.
To ensure that node attribute encoding is independent of sensitive attributes, FairGT partitions the original graph $\mathcal{G}$ into two subgraphs, $\mathcal{G}_0$ and $\mathcal{G}_1$, based on the sensitive attribute values. 
Specifically, $\mathcal{G}_0 = \{v_i \in \mathcal{G} \mid \mathbf{H}[i,s] = 0\}$ and $\mathcal{G}_1 = \{v_i \in \mathcal{G} \mid \mathbf{H}[i,s] = 1\}$, where each subgraph forms a complete graph.
FairGT then constructs a sensitive attribute complete graph $\mathcal{G}_s$ by taking the union of these two subgraphs: $\mathcal{G}_s = \mathcal{G}_0 \cup \mathcal{G}_1$. 
Let $\mathbf{A}_s$ denote the adjacency matrix of $\mathcal{G}_s$ and $\mathbf{H}$ represent the node attribute matrix. 
The multiplication $\mathbf{A}_s \mathbf{H}$ aggregates 1-hop sensitive information, and repeated multiplication captures higher-order propagation. 
For example, two-hop sensitive information can be computed as $\mathbf{A}_s (\mathbf{A}_s \mathbf{H})$.
The $k$-hop sensitive information matrix is formally defined as:

\begin{equation}
    \label{equ:sensitive matrix}
    \mathbf{H}^{(k)} = \mathbf{A}_s^k \mathbf{H}.
\end{equation}

\par To capture essential topological patterns, the structure matrix $\mathbf{S} \in \mathbb{R}^{n \times t}$ is constructed by selecting the top-$t$ eigenvectors associated with the largest eigenvalues.
To ensure the simultaneous preservation of nodal features and graph topology, the original attribute matrix $\mathbf{H}$ is integrated with the structure matrix $\mathbf{S}$ via a concatenation operation:
\begin{equation}
    \label{equ:structural information}
\mathbf{H}' = \mathbf{H} \parallel \mathbf{S},
\end{equation}
where $\parallel$ represents the concatenation operator, and $\mathbf{H}' \in \mathbb{R}^{n \times (d+t)}$ denotes the resulting fused representation matrix.
These fairness-aware encodings are seamlessly incorporated into the Transformer framework, enabling their application to downstream tasks.
The design of FairGT is shown in Algorithm 1. 

\begin{algorithm}[t]
\caption{Fairness-aware Graph2Token Process}
\begin{algorithmic}[1]
\REQUIRE Graph $\mathcal{G} = (\mathcal{V}, \mathbf{A}, \mathbf{X})$, Tokenizer $\texttt{Tokenizer} \in \{\texttt{hop}, \texttt{spectral}, \texttt{subgraph}, etc.\}$
\ENSURE Token sequence $\mathbf{T}$

\FOR{each node $v_i \in \mathcal{V}$}
    \STATE $\mathbf{x}_i \leftarrow \mathbf{H}[i]$
    \STATE $\mathbf{x}'_i \leftarrow \text{FairnessProcess}(\mathbf{x}_i, \mathbf{A})$
    \STATE  $\mathbf{z}_i \leftarrow \text{Projection}(\texttt{Tokenizer}(\mathbf{x}'_i, \mathbf{A}))$
    \STATE $\mathbf{t}_i \leftarrow \text{Fuse}([\mathbf{x}'_i \| \mathbf{z}_i ])$ 
    \STATE Append $\mathbf{t}_i$ to $\mathbf{T}$
\ENDFOR

\RETURN $\mathbf{T}$
\end{algorithmic}
\end{algorithm}

\par Building on this work, a scalable fairness-aware Graph Transformer, referred to as FairGP, is proposed by~\cite{luo2025fairgp} to support large-scale graphs. 
Graph partitioning is leveraged by FairGP to minimise the influence of higher-order nodes, which often contribute to biases.
To mitigate bias in large-scale Graph Transformers, FairGP employs METIS~\cite{karypis1998a} to partition the input graph into $c$ disjoint clusters. 
Let $\mathcal{G}_p = \{ \mathcal{V}_p, \mathcal{E}_p \}$ represent a subgraph of $\mathcal{G}$, such that the collection of subgraphs satisfies the properties of a complete partition: $\bigcup_{p=1}^c \mathcal{G}_p = \mathcal{G}$ and $\bigcap_{p=1}^c \mathcal{G}_p = \emptyset$.

\par The core innovation of FairGP lies in its optimization of global attention mechanisms to address inherent fairness discrepancies. 
By denoting the original attention matrix as $\hat{\mathbf{A}}$, where $\hat{\mathbf{A}}[v_i, v_j]$ signifies the attention weight between nodes $v_i$ and $v_j$, FairGP effectively decouples the attention into intra-cluster and inter-cluster components. 
Specifically, for node pairs within the same cluster ($\mathcal{V}_p$), the refined attention $\hat{\mathbf{A}}'$ retains the original scores:
\begin{equation}
    \hat{\mathbf{A}}'[v_i, v_j] = \hat{\mathbf{A}}[v_i, v_j], \quad \forall v_i, v_j \in \mathcal{V}_p.
\end{equation}

\par In contrast, following theoretical insights regarding fairness enhancement, the attention between disparate clusters is suppressed to zero:
\begin{equation}
    \hat{\mathbf{A}}'[v_i, v_j] = 0, \quad \forall v_i \in \mathcal{V}_p, v_j \in \mathcal{V}_q \quad (p \neq q).
\end{equation}

\par To further assess the computational efficiency of fairness-aware Graph2Token frameworks, we conduct a rigorous complexity analysis. 
Since Graph Transformers typically incur higher algorithmic overhead than conventional GNNs due to global attention, we compare the execution costs of three representative fairness-aware models against several non-fairness baselines. 
For a consistent evaluation, we standardize the experimental environment by setting the hidden dimensionality to $128$ and utilizing a single-layer, single-head configuration. 
All models are benchmarked over $100$ epochs, with runtime recorded in seconds. 
Notably, the computational latency remains invariant across different sensitive attributes within the same dataset.
This comparison aims to illustrate the trade-off between fairness guarantees and computational overhead. 
The detailed complexity formulas and specific data comparisons are presented in Table~\ref{tab:runtime}.

\begin{table}[t]
    \centering
    \caption{Comparison of runtime performance (in seconds) across methods on multiple datasets.}
    \renewcommand{\arraystretch}{1.25}
    \begin{tabular}{lrrr}
    \toprule
     &\textbf{Pokec-z} & \textbf{Pokec-n} & \textbf{AMiner-L} \\
    \midrule
    DIFFormer (ICLR $2023$) & $138.73$ & $124.43$ & $830.44$\\
    SGFormer (NeurIPS $2023$) & $80.60$ & $79.39$ & $708.34$\\
    Polynormer (ICLR $2024$) & $18.34$ & $15.69$ & $70.03$\\
    CoBFormer (ICML $2024$) & $15.66$ & $11.38$ & $66.84$\\
    FairGT (IJCAI $2024$) & $39.16$ & $37.81$ & $204.68$\\
    FairGP (AAAI $2025$) & $6.88$ & $6.14$ & $32.77$\\
    FairGE (WWW $2026$) & $18.94$ & $18.13$ & $76.34$\\
    \bottomrule
    \end{tabular}
    \label{tab:runtime}
\end{table}

\subsection{Dual Fairness-aware Methods}
\par FL has emerged as a popular paradigm for decentralised graph model training, particularly valued for its ability to ensure data privacy. 
However, ensuring fairness across agents remains a significant challenge, as biases from local data can propagate to the global model.

\par This issue is addressed by~\cite{cui2023equipping} through the introduction of fairness-aware local and global model updates, which utilise a dual-end aggregation strategy to mitigate both data and algorithmic biases. 
Group fairness is enhanced by this method without compromising model accuracy.
The fairness penalty is added to the local model's loss function, combining utility loss (such as entropy loss for node classification) with a fairness penalty term. The loss function is defined as:
\begin{equation}
    \mathcal{L}\left(\omega_t^i ; \mathbf{Z}_i\right)=\mathcal{L}_{\text{util}_t}^i+\alpha \mathcal{L}_{\text {fair}_t}^i,
\end{equation}
where $\mathcal{L}_{\mathrm{util}_t}^i$ represents the utility loss (like entropy loss for node classification) and $\alpha$ is an adjustable hyper-parameter.
To address fairness on the server side, F$^2$GNN introduces a fairness weight vector, $\gamma_{\mathcal{F} t}^{\prime}$, which aggregates multiple group fairness metrics for each client. These metrics, such as statistical parity and equal opportunity, are summed for each client to create a fairness vector:
\begin{equation}
    \gamma_{\mathcal{F} t}^{\prime}=\left[\Delta_{\mathrm{SP}_t}^1+\Delta_{\mathrm{EO}_t}^1, \ldots, \Delta_{\mathrm{SP}_t}^{K^{\prime}}+\Delta_{\mathrm{EO}_t}^{K^{\prime}}\right].
\end{equation}

\par This dual approach ensures that both local and global fairness are considered, balancing utility and fairness in the training process.
Building on this, federated personalised graph-based recommendation systems are designed by~\cite{agrawal2024no} to incorporate fairness considerations alongside differential privacy techniques, effectively mitigating group bias while preserving user privacy. 
Both studies highlight the critical role of fairness during the aggregation process at both client and server ends.

\par In addition, incentive mechanisms to ensure equitable participation in FL are proposed by~\cite{pan2024towards}. 
One such mechanism, specifically designed for graph federated learning, adjusts agent contributions based on model gradient alignment and graph diversity, offering fair compensation for participation. 
Similarly, reinforcement learning is introduced on the server side by~\cite{xia2024enhancing} to optimise aggregation by learning collaboration graphs and balancing fairness and performance in the global model.

\par Collectively, these methods demonstrate significant improvements in fairness and performance, underscoring the effectiveness of dual-end fairness mechanisms and incentive strategies in federated graph learning. 
These advancements highlight the growing emphasis on integrating fairness, privacy, and performance optimisation in decentralised learning frameworks.

\subsection{Dynamic Fairness-aware Methods}
\par While fairness techniques for static graph embeddings have seen notable success, extending these methods to dynamic graphs presents significant challenges. 
One of the main difficulties lies in adapting fairness methods designed for static graphs to the dynamic nature of evolving graphs. Retraining structurally fair static graph embedding models from scratch at each time step does not effectively address the biases introduced by the changing graph structure. 
As the graph evolves, new samples are added, and the relationships between nodes change, leading to the amplification of previously unnoticed biases. This necessitates the development of dynamic fairness-aware methods that account for both inherent biases in static embeddings and the continuous structural changes in dynamic graphs. 
These methods must adapt to evolving biases introduced as new data is incorporated, ensuring fairness is maintained throughout the graph’s evolution.

\par Additionally, dynamic shifts in graph structures require special attention to node behaviour. 
For instance, low-degree vertices may form new connections and transition into higher-degree vertices, while most tail vertices exhibit only slight changes. 
Neglecting these dynamic shifts can degrade overall performance. Treating head vertices and transitioning vertices (from tail to head) in the same way as largely unchanged long-tail vertices often leads to performance declines. 
The former requires more nuanced handling, as they are more sensitive to changes in the graph’s structure, highlighting the importance of tailored methods for these evolving scenarios.

\par To address these challenges, FairDGE, the first structurally fair dynamic graph embedding method, is introduced by~\cite{li2024toward}. 
FairDGE mitigates biased structural changes by jointly embedding connection dynamics and the long-term evolutionary trends of vertex degrees. 
Furthermore, it incorporates a contrastive debiasing technique that applies tailored strategies for distinct types of biased structural evolutions, specifically T2H and Starting-from-Head (SfH) patterns.
\begin{equation}
\mathcal{L}_{\text{Downstream-fair}}=\left\|\mathcal{L}_{D S}\left(H_V^\text{T2H}\right)-\mathcal{L}_{D S}\left(H_{\mathcal{V}}^\text{SfH}\right)\right\|_2,
\end{equation}
where $H_V^\text{T2H}$ and $H_V^\text{SfH}$ are the final learned embeddings of T2H and SfH vertices, respectively.
$\| \cdot \|_2$ denotes the $l_2$ norm, and $\mathcal{L}_{DS}$ is a loss function of the downstream tasks.

This method overcomes the effectiveness bottlenecks commonly observed in dynamic graph embeddings, producing fairer embeddings while maintaining robust overall performance. 
These advancements mark a significant step toward achieving fairness in dynamic graph learning without compromising model effectiveness.

\subsection{Fairness-aware Condensation Methods}
\par Training GNNs on large-scale graphs can be computationally expensive, prompting the rise of graph condensation techniques, which aim to reduce graph size while retaining most of the original knowledge. 
However, existing condensation methods often overlook fairness concerns, particularly in how different node subgroups are treated during compression. 
When GNNs are trained on condensed graphs, they tend to exhibit greater bias compared to those trained on full graphs. 
This bias arises because condensed graphs typically include synthetic nodes that lack explicit group identifiers or sensitive attribute information, making it challenging for traditional fairness-aware methods to function effectively.

\par To mitigate these disparities, Mao et al.~\cite{mao2023gcare} introduced Graph Condensation with Adversarial Regularization (GCARe). 
Unlike traditional condensation methods, GCARe incorporates an adversarial objective directly into the distillation process to ensure equitable treatment across various node subgroups. 
The core of this approach lies in a minimax optimization framework:
\begin{equation}
\begin{aligned}
    &\max \theta \min f \mathcal{L} := \\&\quad-\sum{v \in X} \sum{k=1}^K \mathbb{I}_{{s_v=k}} \log f(\mathcal{G}_v, x_v)k,
\end{aligned}
\end{equation}
where $\mathbb{I}{{\cdot}}$ denotes the indicator function, $f(\cdot)$ represents a vector of predictive probabilities for each subgroup, and $f(\cdot)_k$ specifically denotes the probability associated with the $k$-th subgroup out of $K$ total groups.
By embedding fairness constraints directly into the condensation phase, GCARe effectively prevents the inheritance or amplification of biases in the synthesized, compressed graphs. 
This adversarial mechanism ensures balanced representation across subgroups, thereby enhancing the overall integrity and ethical alignment of the distilled graph representations.

\par Building on these advancements, a novel Fair Graph Distillation (FGD) framework is introduced by~\cite{feng2023fair}. 
The FGD framework employs a bi-level optimisation strategy that explicitly incorporates fairness considerations during graph distillation. 
Fairness disparities across different node subgroups are minimised while preserving predictive performance, providing a balanced solution for training GNNs on smaller, distilled graphs. 
Compared to traditional condensation methods, fairness is significantly enhanced in graph distillation by FGD, offering a more equitable and effective approach for scalable graph representation learning.

\subsection{Fairness-aware Graph Unlearning}
\par In the context of graph unlearning, fairness-aware methods go beyond merely considering node attributes; they also account for the unlearning of edges. 
Given a graph $\mathcal{G} = (\mathcal{V}, \mathcal{E}, \mathbf{H})$ and a fully trained GNN model $g_\omega$, the objective of fairness-aware graph unlearning is to remove each edge $e_{ij} \in \mathcal{E}_f$ from $g_\omega$, where $\mathcal{E}_f$ denotes the set of edges to be forgotten, while mitigating any bias introduced by this removal process.

\par Note that node removal can be interpreted as the simultaneous removal of all edges connected to the target node. 
Both the graph structure $\mathcal{E}$ and the GNN $g_\omega$ jointly determine the node embeddings $\mathbf{Z}$ and the predicted probability for an edge $e_{ij}$, defined as $\mathbf{P}_{u,v} = \mathbf{z}_u^\top \mathbf{z}_v$. 
These embeddings influence not only edge prediction outcomes but also the fairness of learned representations. 
A growing body of literature confirms the significant impact of graph topology on node representations~\cite{li2021on,wang2022improving}.

\par Therefore, fairness-aware graph unlearning aims to simultaneously learn an updated model $g_u$ and an optimal graph structure $\mathbf{A}^*$ such that both forgetting and fairness objectives are satisfied. 
Formally, the problem can be formulated as:
\begin{equation}
g_u, \mathbf{A}^* = \operatorname*{arg}\min_{g, \hat{\mathbf{A}}} \mathcal{L}_{\text{un}}(g, \hat{\mathbf{A}}, \mathbf{H}) + \alpha \mathcal{L}_{\text{fair}}(g, \hat{\mathbf{A}}, \mathbf{H}),
\label{equ:fgu}
\end{equation}
where $\mathcal{L}_{\text{un}}$ is the unlearning loss that reduces the model's retention of information from $\mathcal{E}_f$ while preserving performance on the retained edge set $\mathcal{E}r$. 
This framework is compatible with any graph-based unlearning method, enabling plug-and-play integration of any differentiable unlearning objective. 
The second term, $\mathcal{L}_{\text{fair}}$, penalizes violations of representation fairness. 
The hyperparameter $\alpha$ balances the trade-off between unlearning performance and fairness preservation.

\begin{figure}[t]
    \centering
	\includegraphics[width=0.48\textwidth]{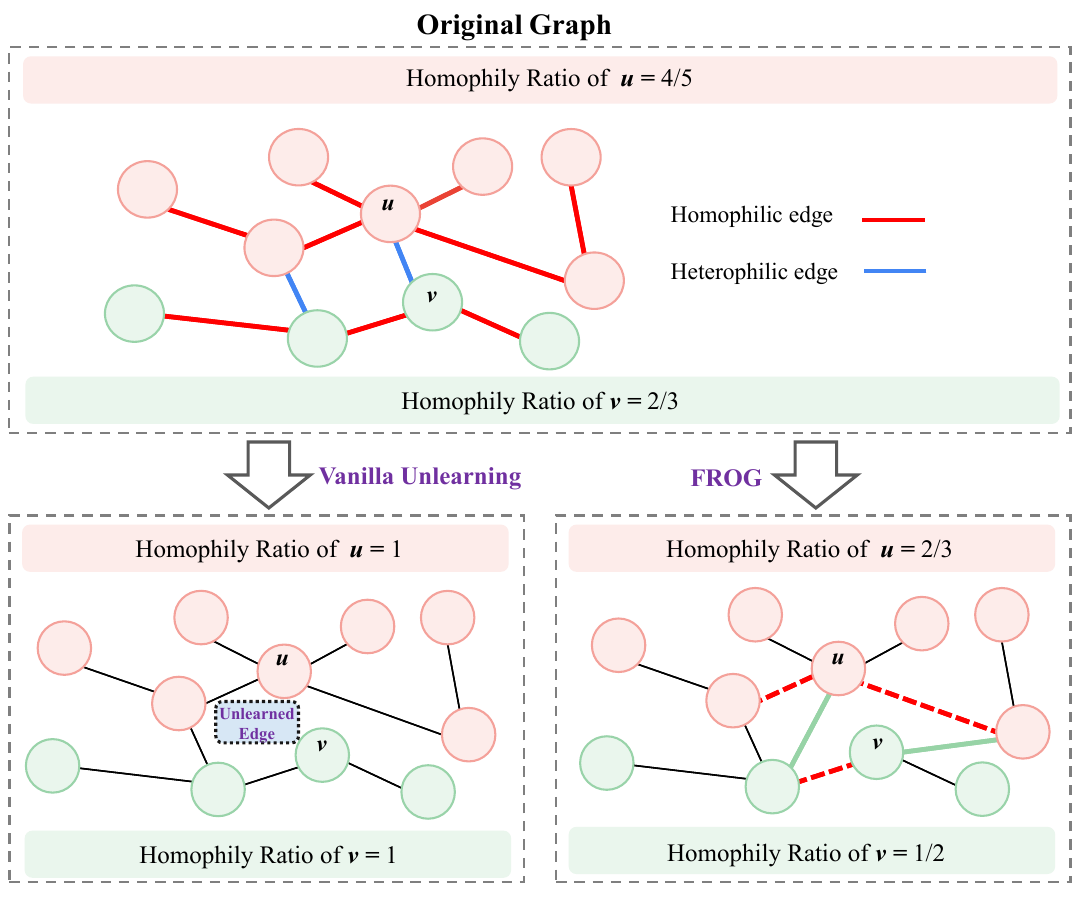}
    \caption{An illustration of FROG~\cite{chen2025frog}.}
    \label{fig:frog}
\end{figure}

\par To prevent excessive deviation from the original graph structure $\mathbf{A}$, fairness-aware graph unlearning also constrains the number of modified edges. 
Specifically, it imposes a limit of at most $N$ changes to the graph, quantified using the $L_0$ norm between $\mathbf{A}$ and $\mathbf{A}^*$. 
The optimization problem thus becomes:
\begin{equation}
    \begin{aligned}
    & g_u, \mathbf{A}^* = \operatorname*{arg}\min_{g, \hat{\mathbf{A}}} \mathcal{L}_{\text{un}}(g, \hat{\mathbf{A}}, \mathbf{H}) + \alpha \mathcal{L}_{\text{fair}}(g, \hat{\mathbf{A}}, \mathbf{H}), \\
    & \text{subject to} \quad |\mathbf{A} - \mathbf{A}^*|_0 \leq N.
    \end{aligned}
\end{equation}

\par Chen et al.~\cite{chen2025frog} proposed a novel approach called Fair Removal on Graphs (FROG), which jointly optimizes the graph structure and the model for fairness-aware unlearning.
The illustration of FROG is shown in Fig.~\ref{fig:frog}.
Specifically, FROG rewires the graph to enhance unlearning efficiency by eliminating redundant edges that obstruct the forgetting process, while simultaneously introducing carefully selected edges to maintain representation fairness. 
Moreover, FROG incorporates a worst-case evaluation mechanism to assess the reliability of the unlearning process under adversarial conditions. 
Extensive experiments on real-world datasets demonstrate the effectiveness of FROG in achieving high unlearning performance without sacrificing fairness.

\section{Open Challenges and Implications}
\par Despite significant advancements in fairness for graph learning, numerous challenges remain, particularly when integrating ML components. 
These challenges highlight the need for novel methods, targeted fairness-aware methods, and foundational frameworks to advance fairness in AGL.

\subsection{Novel Metrics for Augmented Graph Learning}
\par Current research on fairness in graph learning largely relies on traditional fairness metrics, such as statistical parity and equal opportunity for group fairness, or node degree bias. 
However, these metrics are often insufficient for AGL scenarios. 
For instance, in graph federated learning, evaluating fairness necessitates dual-end fairness, i.e., accounting for both client-side and server-side contributions. 
These results cannot be effectively captured through simple linear fitting, requiring separate and specialised metrics for both sides. 
Similarly, in dynamic graph learning, fairness evaluation is complicated by the presence of multiple time steps, where using aggregate metrics like mean or median can skew fairness assessments. 

\par Moreover, traditional metrics often fail to capture item-level bias or user-specific difficulty, especially in graph-based recommender systems. 
Recent work has proposed applying Item Response Theory (IRT) to model the interaction between user ability and item difficulty, enabling a more fine-grained and personalized assessment of fairness and performance~\cite{xu2025towards,xu2025fairness}. 
These IRT-based metrics provide a probabilistic foundation to evaluate recommendation models not just in terms of prediction accuracy, but also in terms of their fairness across subpopulations and item groups.

\par These examples underscore the need for novel, specialized metrics that accurately capture the unique complexities of fairness in AGL settings and downstream applications like recommendation.

\subsection{Enhancing AGL Fairness with Privacy Requirements}
\par One of the critical challenges in AGL lies in its dependence on complete and accurate sensitive attribute information, whether for fairness-aware encoding methods or federated fairness optimization.
However, in real-world scenarios, sensitive attributes (e.g., gender, race, or income level) are frequently missing or intentionally withheld due to privacy regulations or user confidentiality concerns.
This makes it particularly difficult to ensure fairness, as most existing AGL methods require these attributes either for model training or for post-hoc evaluation.

\par A natural workaround involves generating or inferring the missing sensitive attributes through auxiliary models or data augmentation.
Nevertheless, such approaches carry the inherent risk of compromising user privacy, as reconstructing or approximating sensitive data may inadvertently reveal protected information.
Therefore, a crucial open challenge is developing fairness-aware AGL frameworks that operate without relying on sensitive attribute reconstruction.
Consequently, there is a growing demand for attribute-agnostic or non-reconstructive fairness frameworks.

\par Recent advances have shifted toward encoding fairness directly through intrinsic graph properties. For instance, FairGE~\cite{luo2026fairge} addresses fairness in incomplete social networks for Graph Transformers without reconstructing missing data.
Specifically, there is currently a lack of quantitative metrics to rigorously evaluate the latent reconstruction risks inherent in such attribute-agnostic frameworks.
As highlighted in our first Open Challenge (Section V.A), existing fairness research primarily focuses on utility-fairness trade-offs, yet fails to provide standardized indicators to measure privacy leakage resilience.

\par FATE~\cite{kang2023deceptive} and ELEGANT~\cite{dong2025certified}, have taken promising steps in this direction by proposing a certifiable fairness defense mechanism for GNNs that is agnostic to model structure and does not require access to sensitive attributes or model retraining.
This highlights a potential research direction, designing plug-and-play fairness modules for AGL that can certify or enhance fairness even in the absence of complete sensitive attribute information.
Such methods would offer a more privacy-preserving and practically deployable solution for fairness in AGL systems.

\subsection{Targeted Fairness-aware Augmented Graph Learning}
\par Current fairness research in graph learning falls short in addressing the distinct challenges posed by AGL, especially when the integration of ML components fundamentally alters conventional graph learning mechanisms. 
For instance, traditional fairness methods based on static message-passing frameworks become inadequate in scenarios such as graph continual learning, where new nodes, edges, or attributes are incrementally introduced over time~\cite{zhang2022hierarchical, zhang2024continual}. 
In such settings, each newly added sample not only influences the fairness of subsequent predictions but may also exacerbate existing biases or introduce new ones.

\par Addressing fairness under these dynamic conditions necessitates the development of adaptive fairness mechanisms that can respond to temporal shifts in data distribution and graph structure while maintaining fairness guarantees. 
Moreover, as AGL encompasses a wide array of ML paradigms, each introducing different forms of representation, optimization, and inference, there is a critical need for fairness-aware methods that are specifically tailored to the characteristics of each augmentation technique. 
Advancing fairness in AGL therefore requires not only reactive but also proactive design of fairness interventions that align with the evolving nature of graph-based learning environments.

\subsection{The Absence of a Unified FairGX Framework}
\par Although a number of FairGX methods have been developed for AGL, the heterogeneity of underlying ML techniques, ranging from federated optimization to graph transformers and dynamic embedding, necessitates the use of scenario-specific solutions. 
This diversity greatly amplifies the design complexity and implementation cost of fairness interventions, particularly in terms of human labor and computational overhead. 
As AGL continues to evolve across diverse application domains, the lack of generalizable fairness techniques has become a significant bottleneck.

\par While recent efforts have explored graph foundation models, these primarily aim to unify tasks across different granularity levels, such as node-level, edge-level, and graph-level learning~\cite{liu2025graph}. 
However, their focus remains on task generalization, and they rarely address fairness considerations. 
In particular, foundation models tailored for FairGX remain largely unexplored.
Therefore, there is a pressing need to develop a unified FairGX framework that is broadly applicable across AGL paradigms. 
Such methods would enable scalable, reusable, and plug-and-play fairness interventions, significantly reducing human and computational costs. 
Overcoming this challenge is essential to advancing fairness in AGL, equitable, efficient, and sustainable deployment in real-world applications.

\subsection{Fairness-aware LLM4Graph/Graph4LLM}
\par With the growing integration of large language models (LLMs), two emerging paradigms (LLM4Graph and Graph4LLM) have gained significant attention in graph learning~\cite{yu2025graph2text}. 
In LLM4Graph, LLMs are leveraged to enhance graph tasks such as node classification or link prediction via prompt engineering and semantic guidance. 
In contrast, Graph4LLM utilizes graph structures to support LLM tasks, such as reasoning over knowledge graphs or improving retrieval-augmented generation. 
Despite their potential, these methods introduce novel fairness concerns that differ from those in traditional message-passing GNNs.

\par One key challenge arises from the sensitivity of LLMs to prompt phrasing. 
For example, prompts containing gendered terms like waiter or waitress can introduce unintended bias into downstream graph representations or predictions~\cite{chhikara2024few}. 
Such linguistic biases may propagate into LLM4Graph pipelines, resulting in skewed representations or unfair treatment across demographic groups. 
Similarly, in Graph4LLM, the integration of graph structures does not inherently neutralize the bias inherited from pretrained language models, which may amplify disparities in generation quality or content representation.

\par Moreover, conventional fairness-enhancing techniques developed for GNNs, such as adversarial training or message-passing regularization, are not directly applicable in LLM-driven architectures due to their non-localized, prompt-based nature. 
To address this, recent advances in causal mediation analysis offer promising tools to disentangle and quantify the contributions of different components, such as prompt phrasing, graph structure, and model outputs, to fairness-relevant outcomes~\cite{xu2023disentangled2}. 
However, a critical research gap remains: existing causal-based interventions focus almost exclusively on ``pure LLM fairness" (e.g., social bias in text generation) rather than the unique intersectional fairness inherent in LLM4Graph or Graph4LLM paradigms~\cite{yu2025bridging}.

\par Specifically, while Knowledge Graph-Augmented Training (KGAT) has shown potential in mitigating biases by injecting structured, domain-specific knowledge to correct biased associations, its scalability within hybrid LLM4Graph/Graph4LLM pipelines remains unrefined~\cite{kumar2025detecting}. 
Furthermore, although causal analysis can estimate and filter out unaligned or ``negative transfer" data to ensure positive fairness transfer (as seen in Fair-Gender approaches), these methods have yet to account for the dynamic topology of graphs. 
In LLM4Graph/Graph4LLM systems, the ``alignment" required for fairness is not merely semantic but structural; the model must reconcile attribute-balanced natural language with the non-i.i.d. dependencies of the graph. 

\par Consequently, fairness-aware adaptations that jointly consider graph topology, natural language prompts, and LLM representation bias remain largely underexplored. 
Bridging this gap represents a vital and uncharted research direction for developing robust, fairness-aware LLM4Graph/Graph4LLM integrated systems.

\section{Conclusion}
\par In conclusion, while graph learning has demonstrated effectiveness in various real-world applications, the augmentation of ML into graph learning introduces new fairness challenges. 
Traditional fairness-aware methods often address biases in message passing or node attributes, but AGL adds complexity with additional sources of bias, such as those arising from federated learning or attention mechanisms in graph transformers. 
As these techniques gain traction in high-risk domains, addressing the fairness issues they introduce is essential to prevent discriminatory outcomes. 
This review underscores the need for novel fairness strategies tailored to AGL, ensuring that these powerful techniques promote equity and ethical decision-making in applications like job recommendations and criminal justice.

\par Looking ahead, we identify five key areas for future research:
Developing dual-end and IRT-based metrics to capture complex biases in federated and recommendation scenarios.
Designing plug-and-play modules that ensure fairness without requiring access to sensitive user attributes.
Moving toward generalizable, fairness-aware foundation models that can adapt to continual and dynamic graph shifts.
Investigating how prompt-based biases and semantic guidance affect fairness in LLM4Graph and Graph4LLM paradigms.
By addressing these frontiers, the field can ensure that the next generation of graph learning systems is not only powerful but also equitable and ethically sound.

\bibliographystyle{IEEEtran}
\bibliography{IEEEabrv,ref}

\begin{thebibliography}{100}
\providecommand{\url}[1]{#1}
\csname url@samestyle\endcsname
\providecommand{\newblock}{\relax}
\providecommand{\bibinfo}[2]{#2}
\providecommand{\BIBentrySTDinterwordspacing}{\spaceskip=0pt\relax}
\providecommand{\BIBentryALTinterwordstretchfactor}{4}
\providecommand{\BIBentryALTinterwordspacing}{\spaceskip=\fontdimen2\font plus
\BIBentryALTinterwordstretchfactor\fontdimen3\font minus \fontdimen4\font\relax}
\providecommand{\BIBforeignlanguage}[2]{{%
\expandafter\ifx\csname l@#1\endcsname\relax
\typeout{** WARNING: IEEEtran.bst: No hyphenation pattern has been}%
\typeout{** loaded for the language `#1'. Using the pattern for}%
\typeout{** the default language instead.}%
\else
\language=\csname l@#1\endcsname
\fi
#2}}
\providecommand{\BIBdecl}{\relax}
\BIBdecl

\bibitem{xia2021graph}
F.~Xia, K.~Sun, S.~Yu, A.~Aziz, L.~Wan, S.~Pan, and H.~Liu, ``Graph learning: A survey,'' \emph{IEEE Transactions on Artificial Intelligence}, vol.~2, no.~2, pp. 109--127, 2021.

\bibitem{guo2023contranorm}
X.~Guo, Y.~Wang, T.~Du, and Y.~Wang, ``Contra{N}orm: A contrastive learning perspective on oversmoothing and beyond,'' in \emph{International Conference on Learning Representations}, 2023.

\bibitem{he2023generalization}
X.~He, B.~Hooi, T.~Laurent, A.~Perold, Y.~LeCun, and X.~Bresson, ``A generalization of {V}it/{M}lp-mixer to graphs,'' in \emph{Proceedings of the 40th International Conference on Machine Learning}, 2023, pp. 12\,724--12\,745.

\bibitem{gao2024graph}
X.~Gao, T.~Chen, Y.~Zang, W.~Zhang, Q.~V.~H. Nguyen, K.~Zheng, and H.~Yin, ``Graph condensation for inductive node representation learning,'' in \emph{2024 IEEE 40th International Conference on Data Engineering}, 2024.

\bibitem{tang2024personlized}
T.~Tang, Z.~Han, Z.~Cai, S.~Yu, X.~Zhou, T.~Oseni, and S.~K. Das, ``Personalized federated graph learning on {N}on-{IID} electronic health records,'' \emph{IEEE Transactions on Neural Networks and Learning Systems}, vol.~35, pp. 11\,843--11\,856, 2024.

\bibitem{pei2024memory}
H.~Pei, Y.~Xiong, P.~Wang, J.~Tao, J.~Liu, H.~Deng, J.~Ma, and X.~Guan, ``Memory disagreement: A {P}seudo-labeling measure from training dynamics for semi-supervised graph learning,'' in \emph{Proceedings of the ACM Web Conference 2024}, 2024, pp. 434--445.

\bibitem{traag2019louvain}
V.~A. Traag, L.~Waltman, and N.~J. Van~Eck, ``From louvain to leiden: Guaranteeing well-connected communities,'' \emph{Scientific Reports}, vol.~9, no.~1, pp. 1--12, 2019.

\bibitem{wan2024federated}
G.~Wan, W.~Huang, and M.~Ye, ``Federated graph learning under domain shift with generalizable prototypes,'' in \emph{Proceedings of the AAAI Conference on Artificial Intelligence}, 2024, pp. 15\,429--15\,437.

\bibitem{chen2024easydgl}
C.~Chen, H.~Geng, N.~Yang, X.~Yang, and J.~Yan, ``Easy{DGL}: Encode, train and interpret for continuous-time dynamic graph learning,'' \emph{IEEE Transactions on Pattern Analysis and Machine Intelligence}, vol.~46, no.~12, pp. 10\,845--10\,862, 2024.

\bibitem{wu2023demystifying}
X.~Wu, A.~Ajorlou, Z.~Wu, and A.~Jadbabaie, ``Demystifying oversmoothing in attention-based graph neural networks,'' in \emph{Proceedings of the 36th Conference on Neural Information Processing Systems}, 2023, pp. 35\,084--35\,106.

\bibitem{jin2022graph}
W.~Jin, L.~Zhao, S.~Zhang, Y.~Liu, J.~Tang, and N.~Shah, ``Graph condensation for graph neural networks,'' in \emph{International Conference on Learning Representations}, 2022.

\bibitem{modak2024cpa}
S.~Modak, A.~Malhotra, S.~Malik, A.~Surisetty, and E.~Abdel-Raheem, ``{CP}a-{WAC}: Constellation partitioning-based scalable weighted aggregation composition for knowledge graph embedding,'' in \emph{Proceedings of the 32rd International Joint Conference on Artificial Intelligence}, 2024, pp. 3504--3512.

\bibitem{luo2025fairness}
Y.~Luo, Z.~Li, Q.~Liu, and J.~Zhu, ``Fairness without demographic through learning graph of gradients,'' in \emph{Proceedings of the 31th ACM SIGKDD Conference on Knowledge Discovery and Data Mining}, 2025, pp. 918--926.

\bibitem{chen2023bias}
J.~Chen, H.~Dong, X.~Wang, F.~Feng, M.~Wang, and X.~He, ``Bias and debias in recommender system: A survey and future directions,'' \emph{ACM Transactions on Information Systems}, vol.~41, no.~3, pp. 1--39, 2023.

\bibitem{xia2023cengcn}
F.~Xia, L.~Wang, T.~Tang, X.~Chen, X.~Kong, G.~Oatley, and I.~King, ``Cen{GCN}: Centralized convolutional networks with vertex imbalance for scale-free graphs,'' \emph{IEEE Transactions on Knowledge and Data Engineering}, vol.~35, no.~5, pp. 4555--4569, 2023.

\bibitem{feng2023criminal}
G.~Feng, Y.~Qin, R.~Huang, and Y.~Chen, ``Criminal action graph: A semantic representation model of judgement documents for legal charge prediction,'' \emph{Information Processing \& Management}, vol.~60, no.~5, p. 103421, 2023.

\bibitem{cheng2023regulating}
D.~Cheng, Z.~Niu, J.~Li, and C.~Jiang, ``Regulating systemic crises: Stemming the contagion risk in networked-loans through deep graph learning,'' \emph{IEEE Transactions on Knowledge and Data Engineering}, vol.~35, no.~6, pp. 6278--6289, 2023.

\bibitem{kang2022algorithmic}
J.~Kang and H.~Tong, ``Algorithmic fairness on graphs: Methods and trends,'' in \emph{Proceedings of the 28th ACM SIGKDD Conference on Knowledge Discovery and Data Mining}, 2022, pp. 4798--4799.

\bibitem{dong2023fairness}
Y.~Dong, J.~Ma, S.~Wang, C.~Chen, and J.~Li, ``Fairness in graph mining: A survey,'' \emph{IEEE Transactions on Knowledge and Data Engineering}, vol.~35, no.~10, pp. 10\,583--10\,602, 2023.

\bibitem{dai2024comprehensive}
E.~Dai, T.~Zhao, H.~Zhu, J.~Xu, Z.~Guo, H.~Liu, J.~Tang, and S.~Wang, ``A comprehensive survey on trustworthy graph neural networks: Privacy, robustness, fairness, and explainability,'' \emph{Machine Intelligence Research}, pp. 1--51, 2024.

\bibitem{purificato2025gnn}
E.~Purificato, H.~J. Mahadik, L.~Boratto, and E.~W.~D. Luca, ``{GNN}'s {FAME}: Fairness-aware messages for graph neural networks,'' in \emph{Proceedings of the 33rd ACM Conference on User Modeling, Adaptation and Personalization}, 2025, pp. 301--306.

\bibitem{dai2023learning}
E.~Dai and S.~Wang, ``Learning fair graph neural networks with limited and private sensitive attribute information,'' \emph{IEEE Transactions on Knowledge and Data Engineering}, vol.~35, no.~7, pp. 7103--7117, 2023.

\bibitem{xu2025fairness}
Z.~Xu, S.~Kandannaarachchi, C.~S. Ong, and E.~Ntoutsi, ``Fairness evaluation with item response theory,'' in \emph{Proceedings of the ACM Web Conference 2025}, 2025, pp. 2276--2288.

\bibitem{li2025fairness}
Z.~Li, M.~Lin, J.~Wang, and S.~Wang, ``Fairness-aware prompt tuning for graph neural networks,'' in \emph{Proceedings of the ACM Web Conference 2025}, 2025, pp. 3586--3597.

\bibitem{ye2026fairgse}
Z.~Ye, J.~Lu, T.~Gu, F.~Hao, and X.~Wang, ``Fair{GSE}: Fairness-aware graph neural network without high false positive rates,'' in \emph{Proceedings of the AAAI Conference on Artificial Intelligence}, 2026.

\bibitem{zhao2025fs}
J.~Zhao, T.~Huang, S.~Liu, J.~Yin, Y.~Pei, M.~Fang, and M.~Pechenizkiy, ``{FS}-{GNN}: Improving fairness in graph neural networks via joint sparsification,'' \emph{Neurocomputing}, vol. 648, p. 130641, 2025.

\bibitem{xie2022fairrankvis}
T.~Xie, Y.~Ma, J.~Kang, H.~Tong, and R.~Maciejewski, ``Fairrankvis: A visual analytics framework for exploring algorithmic fairness in graph mining models,'' \emph{IEEE Transactions on Visualization and Computer Graphics}, vol.~28, no.~1, pp. 368--377, 2022.

\bibitem{chan2024group}
E.~Chan, Z.~Liu, R.~Qiu, Y.~Zhang, R.~Maciejewski, and H.~Tong, ``Group fairness via group consensus,'' in \emph{Proceedings of the 2024 ACM Conference on Fairness, Accountability, and Transparency}, 2024, pp. 1788--1808.

\bibitem{dong2021individual}
Y.~Dong, J.~Kang, H.~Tong, and J.~Li, ``Individual fairness for graph neural networks: A ranking based approach,'' in \emph{Proceedings of the 27th ACM SIGKDD International Conference on Knowledge Discovery and Data Mining}, 2021, pp. 300--310.

\bibitem{he2024on}
X.~He, J.~Kang, R.~Qiu, F.~Wang, J.~Sepulveda, and H.~Tong, ``On the sensitivity of individual fairness: Measures and robust algorithms,'' in \emph{Proceedings of the 33rd ACM International Conference on Information and Knowledge Management}, 2024, pp. 829--838.

\bibitem{zhu2025sagif}
Y.~Zhu, J.~Li, H.~Zhang, L.~Chen, and Z.~Zheng, ``Sa{GIF}: Improving individual fairness in graph neural networks via similarity encoding,'' \emph{arXiv preprint arXiv:2506.18696}, 2025.

\bibitem{dwork2012fairness}
C.~Dwork, M.~Hardt, T.~Pitassi, O.~Reingold, and R.~S. Zemel, ``Fairness through awareness,'' in \emph{Proceedings of the 3rd Innovations in Theoretical Computer Science Conference}, 2012, pp. 214--226.

\bibitem{jiang2023fairness}
Z.~Jiang, X.~Han, C.~Fan, Z.~Liu, A.~Mostafavi, and X.~Hu, ``Fairness and explainability: Bridging the gap towards fair model explanations,'' in \emph{Proceedings of the AAAI Conference on Artificial Intelligence}, 2023.

\bibitem{hardt2016equality}
M.~Hardt, E.~Price, and N.~Srebro, ``Equality of opportunity in supervised learning,'' in \emph{Proceedings of the 30th International Conference on Neural Information Processing Systems}, 2016, pp. 3323--3331.

\bibitem{liu2022fairness}
S.~Liu, Y.~Ge, S.~Xu, Y.~Zhang, and A.~Marian, ``Justifying recommendations using distantly-labeled reviews and fine-grained aspects,'' in \emph{Proceedings of the 16th ACM Conference on Recommender Systems}, 2022, pp. 168--178.

\bibitem{lahoti2019operationalizing}
P.~Lahoti, K.~P. Gummadi, and G.~Weikum, ``Operationalizing individual fairness with pairwise fair representations,'' in \emph{Proceedings of the VLDB Endowment}, 2019, pp. 506--518.

\bibitem{chen2024polygcl}
J.~Chen, R.~Lei, and Z.~Wei, ``{POLYGCL}: Graph contrasstive learning via learnable spectral polonomial filters,'' in \emph{International Conference on Learning Representations}, 2024.

\bibitem{kang2022rawlsgcn}
J.~Kang, Y.~Zhu, Y.~Xia, J.~Luo, and H.~Tong, ``{RawlsGCN}: Towards rawlsian difference principle on graph convolutional network,'' in \emph{Proceedings of the ACM Web Conference 2022}, 2022, pp. 1214--1225.

\bibitem{pan2024towards}
C.~Pan, J.~Xu, Y.~Yu, Z.~Yang, Q.~Wu, C.~Wang, L.~Chen, and Y.~Yang, ``Towards fair graph federated learning via incentive mechanisms,'' in \emph{Proceedings of the AAAI Conference on Artificial Intelligence}, 2024, pp. 14\,499--14\,507.

\bibitem{xu2021gradient}
X.~Xu, L.~Lyu, X.~Ma, C.~Miao, C.~S. Foo, and B.~K.~K. Low, ``Gradient-driven rewards to guarantee fairness in collaborative machine learning,'' in \emph{Proceedings of the 35th International Conference on Neural Information Processing Systems}, 2021, pp. 16\,104--16\,117.

\bibitem{xia2024enhancing}
Y.~Xia, B.~Ma, Q.~Dou, and Y.~Xia, ``Enhancing federated learning performance fairness via collaboration graph-based reinforcement learning,'' in \emph{Proceedings of the Medical Image Computing and Computer Assisted Intervention}, 2024, pp. 263--272.

\bibitem{wang2021federated}
Z.~Wang, X.~Fan, J.~Qi, C.~Wen, C.~Wang, and R.~Yu, ``Federated learning with fair averaging,'' in \emph{Proceedings of the 30th International Joint Conference on Artificial Intelligence}, 2021, pp. 1615--1623.

\bibitem{mao2023gcare}
R.~Mao, W.~Fan, and Q.~Li, ``{GCAR}e: Mitigating subgroup unfairness in graph condensation through adversarial regularization,'' \emph{Applied Science}, vol.~13, no.~16, p. 9166, 2023.

\bibitem{yang2017measuring}
K.~Yang and J.~Stoyanovich, ``Measuring fairness in ranked outpus,'' in \emph{Proceedings of the 29th International Conference on Scientific and Statistical Database Management}, 2017, pp. 1--6.

\bibitem{li2024contrastive}
C.~Li, D.~Cheng, G.~Zhang, and S.~Zhang, ``Contrastive learning for fair graph representations via counterfactual graph augmentation,'' \emph{Knowledge-Based Systems}, vol. 305, p. 112635, 2024.

\bibitem{zhang2025disentangled}
G.~Zhang, G.~Yuan, D.~Cheng, L.~Liu, J.~Li, and S.~Zhang, ``Disentangled contrastive learning for fair graph representations,'' \emph{Neural Networks}, vol. 181, p. 106781, 2025.

\bibitem{zhu2024the}
Y.~Zhu, J.~Li, L.~Chen, and Z.~Zheng, ``The devil is in the data: Learning fair graph neural networks via partial knowledge distillation,'' in \emph{Proceedings of the 17th ACM International Conference on Web Search and Data Mining}, 2024, pp. 1012--1021.

\bibitem{zhu2024fair}
Y.~Zhu, J.~Li, Z.~Zheng, and L.~Chen, ``Fair graph representation learning via sensitive attribute disentanglement,'' in \emph{Proceedings of the ACM Web Conference 2024}, 2024, pp. 1182--1192.

\bibitem{chen2023privacy}
H.~Chen, T.~Zhu, T.~Zhang, W.~Zhou, and P.~S. Yu, ``Privacy and fairness in federated learning: On the perspective of tradeoff,'' \emph{ACM Computing Surveys}, vol.~56, no.~2, pp. 1--37, 2023.

\bibitem{ezzeldin2023fairfed}
Y.~H. Ezzeldin, S.~Yan, C.~He, E.~Ferrara, and A.~S. Avestimehr, ``Fair{F}ed: Enabling group fairness in federated learning,'' in \emph{Proceedings of the AAAI Conference on Artificial Intelligence}, 2023, pp. 7494--7502.

\bibitem{shi2024towards}
Y.~Shi, H.~Yu, and C.~Leung, ``Towards fairness-aware federated learning,'' \emph{IEEE Transactions on Neural Networks and Learning Systems}, vol.~35, no.~9, pp. 11\,922--11\,938, 2024.

\bibitem{luo2026fairge}
R.~Luo, H.~Huang, T.~Tang, J.~Ren, Z.~Xu, M.~Hou, E.~Dai, and F.~Xia, ``Fair{GE}: Fairness-aware graph encoding in incomplete social networks,'' in \emph{Proceedings of the ACM Web Conference 2026}, 2026.

\bibitem{zayed2024fairness}
A.~Zayed, G.~Mordido, S.~Shabanian, L.~Baldini, and S.~Chandar, ``Fairness-aware structured pruning in transformers,'' in \emph{Proceedings of the AAAI Conference on Artificial Intelligence}, 2024, pp. 22\,484--22\,492.

\bibitem{xu2023disentangled}
Z.~Xu, D.~Cheng, J.~Li, J.~Liu, L.~Liu, and K.~Wang, ``Disentangled representation for causal mediation analysis,'' in \emph{Proceedings of the AAAI Conference on Artificial Intelligence}, 2023, pp. 10\,666--10\,674.

\bibitem{xu2022assessing}
Z.~Xu, Z.~Xu, J.~Liu, D.~Cheng, J.~Li, L.~Liu, and K.~Wang, ``Assessing classifier fairness with collider bias,'' in \emph{Pacific-Asia Conference on Knowledge Discovery and Data Mining}, 2022, pp. 262--276.

\bibitem{robertson2024fairpfn}
J.~Robertson, N.~Hollmann, N.~Awad, and F.~Hutter, ``Fair{PFN}: Transformers can do counterfactual fairness,'' \emph{arXiv preprint arXiv:2407.05732}, 2024.

\bibitem{wang2022uncovering}
R.~Wang, X.~Wang, C.~Shi, and L.~Song, ``Uncovering the structural fairness in graph contrastive learning,'' in \emph{Proceedings of the 36th Conference on Neural Information Processing Systems}, 2022, pp. 32\,465--32\,473.

\bibitem{xing2024less}
Y.~Xing, X.~Wang, Y.~Li, H.~Huang, and C.~Shi, ``Less is more: On the over-globalizing problem in graph transformers,'' in \emph{Proceedings of the 41st International Conference on Machine Learning}, 2024, pp. 54\,656--54\,672.

\bibitem{shehzad2024graph}
A.~Shehzad, F.~Xia, S.~Abid, C.~Peng, S.~Yu, D.~Zhang, and K.~Verspoor, ``Graph transformers: A survey,'' \emph{arXiv preprint arXiv:2407.09777}, 2024.

\bibitem{luo2025fairgp}
R.~Luo, H.~Huang, L.~Ivan, C.~Xu, J.~Qi, and F.~Xia, ``Fair{GP}: A scalable and fair graph transformer using graph partitioning,'' in \emph{Proceedings of the AAAI Conference on Artificial Intelligence}, 2025.

\bibitem{feng2023fair}
Q.~Feng, Z.~Jiang, R.~Li, Y.~Wang, N.~Zou, J.~Bian, and X.~Hu, ``Fair graph distillation,'' in \emph{Proceedings of the 37th International Conference on Neural Information Processing Systems}, 2023, pp. 80\,644--80\,660.

\bibitem{cui2023equipping}
N.~Cui, X.~Wang, W.~H. Wang, V.~Chen, and Y.~Ning, ``Equipping federated graph neural networks with structure-aware group fairness,'' in \emph{2023 IEEE International Conference on Data Mining}, 2023, pp. 980--985.

\bibitem{agrawal2024no}
N.~Agrawal, A.~K. Sirohi, S.~Kumar, and Jayadeva, ``No prejudice! fair federated graph neural networks for personalized recommendation,'' in \emph{Proceedings of the AAAI Conference on Artificial Intelligence}, 2024, pp. 10\,775--10\,783.

\bibitem{kang2021fair}
J.~Kang and H.~Tong, ``Fair graph mining,'' in \emph{Proceedings of the 30th ACM International Conference on Information and Knowledge Management}, 2021, pp. 4849--4852.

\bibitem{luo2026utility}
R.~Luo, H.~Huang, S.~Yu, F.~Yu, F.~Xia, S.~K. Das, and C.~Zhang, ``Utility-preserving federated graph learning with dual-perspective fairness,'' \emph{IEEE Transactions on Pattern Analysis and Machine Intelligence}, 2026.

\bibitem{wang2023mitigating}
G.~Wang, A.~Payani, M.~Lee, and R.~Kompella, ``Mitigating group bias in federated learning: Beyond local fairness,'' \emph{arXiv preprint arXiv:2305.09931}, 2023.

\bibitem{harper2015the}
F.~M. Harper and J.~A. Konstan, ``The {M}ovie{L}ens datasets: History and context,'' \emph{ACM Transactions on Interactive Intelligence Systems}, vol.~5, no.~4, pp. 1--19, 2015.

\bibitem{ni2019justifying}
J.~Ni, J.~Li, and J.~Mcauley, ``Justifying recommendations using distantly-labeled reviews and fine-grained aspects,'' in \emph{Proceedings of the 2019 Conference on Empirical Methods in Natural Language Processing and the 9th International Joint Conference on Natural Language Processing}, 2019, pp. 188--197.

\bibitem{li2024toward}
Y.~Li, Y.~Yang, J.~Cao, S.~Liu, H.~Tang, and G.~Xu, ``Toward structure fairness in dynamic graph embedding: A trend-aware dual debiasing approach,'' in \emph{Proceedings of the 30th ACM SIGKDD Conference on Knowledge Discovery and Data Mining}, 2024, pp. 1701--1712.

\bibitem{takac2012data}
L.~Takac and Z.~Michal, ``Data analysis in public social networks,'' in \emph{Proceedings of International Scientific Conference and International Workshop Present Day Trends of Innovations}, 2012.

\bibitem{wan2019aminer}
H.~Wan, Y.~Zhang, J.~Zhang, and J.~Tang, ``{AM}iner: Search and mining of academic social networks,'' \emph{Data Intelligence}, vol.~1, no.~1, pp. 58--76, 2019.

\bibitem{asuncion2007uci}
A.~Asuncion and D.~Newman, ``Uci machine learning repository,'' 2007.

\bibitem{jordan2015effect}
K.~L. Jordan and T.~L. Freiburger, ``The effect of race/ethnicity on sentencing: Examining sentence type, jail length, and prison length,'' \emph{Journal of Ethnicity in Criminal Justice}, vol.~13, no.~3, pp. 179--196, 2015.

\bibitem{yeh2009the}
Y.~I-cheng and L.~Che-hui, ``The comparisons of data mining techniques for the predictive accuracy of probability of default of credit card clients,'' \emph{Expert Systems with Applications}, vol.~36, no.~2, pp. 2473--2480, 2009.

\bibitem{kipf2017semi}
T.~N. Kipf and M.~Welling, ``Semi-supervised classification with graph convolutional networks,'' in \emph{International Conference on Learning Representations}, 2017.

\bibitem{hu2020open}
W.~Hu, M.~Fey, M.~Zitnik, Y.~Dong, H.~Ren, B.~Liu, M.~Catasta, and J.~Leskovec, ``Open graph benchmark: Datasets for machine learning on graphs,'' in \emph{Proceedings of the 34th Conference on Neural Information Processing Systems}, 2020, pp. 22\,118--22\,133.

\bibitem{bojchevski2018deep}
A.~Bojchevski and S.~Guunemann, ``Deep {G}aussian embeddings of graphs: Unsupervised inductive learning via ranking,'' in \emph{International Conference on Learning Representations}, 2018.

\bibitem{leskovec2012learning}
J.~Leskovec and J.~Mcauley, ``Learning to discover social circles in ego networks,'' in \emph{Proceedings of the 25th International Conference on Neural Information Processing Systems}, 2012, pp. 539--547.

\bibitem{borgwardt2005protein}
K.~M. Borgwardt, C.~S. Ong, S.~Schonauer, S.~Vishwanathan, A.~J. Smola, and H.-P. Kriegel, ``Protein function prediction via graph kernels,'' \emph{Bioinformatics}, vol.~21, no.~1, pp. 47--56, 2005.

\bibitem{dobson2003distinguishing}
P.~D. Dobson and A.~J. Doig, ``Distinguishing enzyme structures from non-enzymes without alignments,'' \emph{Journal of Molecular Biology}, vol. 330, no.~4, pp. 771--783, 2003.

\bibitem{yanardag2015deep}
P.~Yanardag and S.~Vishwanathan, ``Deep graph kernels,'' in \emph{Proceedings of the 21th ACM SIGKDD International Conference on Knowledge Discovery and Data Mining}, 2015, pp. 1365--1374.

\bibitem{tschandl2018the}
P.~Tschandl, C.~Rosendahl, and H.~Kittler, ``The {HAM10000} dataset, a large collection of multi-source dermatoscopic images of common pigmented skin lesions,'' \emph{Scientific data}, vol.~5, no.~1, pp. 1--9, 2018.

\bibitem{agarwal2021towards}
C.~Agarwal, H.~Lakkaraju, and M.~Zitnik, ``Towards a unified framework for fair and stable graph representation learning,'' in \emph{Proceedings of the 37th Conference on Uncertainty in Artificial Intelligence}, 2021, pp. 2114--2124.

\bibitem{maggie2019aptos}
K.~Maggie and S.~Dane, ``{APTOS} 2019 blindness detection,'' 2019.

\bibitem{liu2022deepdrid}
R.~Liu, X.~Wang, Q.~Wu, L.~Dai, X.~Fang, T.~Yan, J.~Son, S.~Tang, J.~Li, Z.~Gao \emph{et~al.}, ``Deep{DR}i{D}: Diabetic retinopathy—grading and image quality estimation challenge,'' \emph{Patterns}, vol.~3, no.~6, 2022.

\bibitem{zhou2020benchmark}
Y.~Zhou, B.~Wang, L.~Huang, S.~Cui, and L.~Shao, ``A benchmark for studying diabetic retinopathy: Segmentation, grading, and transferability,'' \emph{IEEE Transactions on Medical Imaging}, vol.~40, no.~3, pp. 818--828, 2020.

\bibitem{decenciere2013teleophta}
E.~Decenciere, G.~Cazuguel, X.~Zhang, G.~Thibault, J.-C. Klein, F.~Meyer, B.~Marcotegui, G.~Quellec, M.~Lamard, R.~Danno, D.~Elie, P.~Massin, Z.~Viktor, A.~Erginay, B.~Lay, and A.~Chabouis, ``Tele{O}phta: Machine learning and image processing methods for teleophthalmology,'' \emph{{IRBM}}, vol.~34, no.~2, pp. 196--203, 2013.

\bibitem{porwal2018indian}
P.~Porwal, S.~Pachade, R.~Kamble, M.~Kokare, G.~Deshmukh, V.~Sahasrabuddhe, and F.~Meriaudeau, ``Indian diabetic retinopathy image dataset (idrid): A database for diabetic retinopathy screening research,'' \emph{Data}, vol.~3, no.~3, p.~25, 2018.

\bibitem{abramoff2013automated}
M.~D. Abramoff, J.~C. Folk, D.~P. Han, J.~Wlaker, D.~F. Williams, S.~R. Russell, P.~Massin, B.~Cochener, C.~Gain, L.~Tang, M.~Lamard, D.~C. Moga, G.~Quellec, and M.~Niemeijer, ``Automated analysis of retinal images for dtection of refereable diabetic retinopathy,'' \emph{JAMA Ophthalmol}, vol. 131, no.~3, pp. 351--357, 2013.

\bibitem{luo2024fairgt}
R.~Luo, H.~Huang, S.~Yu, X.~Zhang, and F.~Xia, ``Fair{GT}: A fairness-aware graph transformer,'' in \emph{Proceedings of the 32rd International Joint Conference on Artificial Intelligence}, 2024, pp. 449--457.

\bibitem{gao2026fairgc}
Y.~Gao, C.~Huang, W.~Shi, K.~Sun, Z.~Xu, X.~Zhang, M.~Hou, and R.~Luo, ``Fair{GC}: Fairness-aware graph condensation,'' in \emph{Proceedings of the 2026 International Joint Conference on Neural Networks}, 2026.

\bibitem{chen2025frog}
Z.~Chen, J.~Chen, G.~Tolomei, S.~Liu, H.~Amiri, Y.~Wang, K.~Nag, and L.~Lin, ``{FROG}: Fair removal on graphs,'' \emph{arXiv preprint arXiv:2503.18197}, 2025.

\bibitem{luo2026fairgu}
R.~Luo, Y.~Yang, H.~Huang, Q.~Qing, M.~Hou, Z.~Xu, Y.~Yu, J.~Zhou, and F.~Xia, ``Fair{GU}: Fairness-aware graph unlearning in social networks,'' in \emph{Proceedings of the ACM Web Conference 2026}, 2026.

\bibitem{wang2026guic}
Z.~Wang, T.~Liu, and W.~Zhang, ``{GUIC}: Certified graph unlearning with individual fairness guarantees,'' in \emph{Proceedings of the AAAI Conference on Artificial Intelligence}, 2026.

\bibitem{karypis1998a}
G.~Karypis and V.~Kumar, ``A fast and high quality multilevel scheme for partitioning irregular graphs,'' \emph{SIAM Journal on Scientific Computing}, vol.~20, no.~1, pp. 359--392, 1998.

\bibitem{li2021on}
P.~Li, Y.~Wang, H.~Zhao, P.~Hong, and H.~Liu, ``On dyadic fairness: Exploring and mitigating bias inn graph connections,'' in \emph{International Conference on Learning Representations}, 2021.

\bibitem{wang2022improving}
Y.~Wang, Y.~Zhao, Y.~Dong, H.~Chen, J.~Li, and T.~Derr, ``Improving fairness in graph neural networks via mitigating sensitive attribute leakage,'' in \emph{Proceedings of the 28th ACM SIGKDD Conference on Knowledge Discovery and Data Mining}, 2022, pp. 1938--1948.

\bibitem{xu2025towards}
Z.~Xu, C.~Ma, Y.~Ren, J.~Chan, W.~Shao, and F.~Xia, ``Towards better evaluation of recommendation algorithms with bi-directional item response theory,'' in \emph{Companion Proceedings of the ACM on Web Conference 2025}, 2025, pp. 1455--1459.

\bibitem{kang2023deceptive}
J.~Kang, Y.~Xia, R.~Maciejewski, J.~Luo, and H.~Tong, ``Deceptive fairness attacks on graphs via meta learning,'' \emph{arXiv preprint arXiv:2310.15653}, 2023.

\bibitem{dong2025certified}
Y.~Dong, B.~Zhang, H.~Tong, and J.~Li, ``Certified defense on the fairness of graph neural networks,'' \emph{arXiv preprint arXiv:2311.02757}, 2025.

\bibitem{zhang2022hierarchical}
X.~Zhang, D.~Song, and D.~Tao, ``Hierarchical prototype networks for continual graph representation learning,'' \emph{IEEE Transactions on Pattern Analysis and Machine Intelligence}, vol.~45, no.~4, pp. 4622--4636, 2022.

\bibitem{zhang2024continual}
------, ``Continual learning on graphs: Challenges, solutions, and opportunities,'' \emph{arXiv preprint arXiv:2402.11565}, 2024.

\bibitem{liu2025graph}
J.~Liu, C.~Yang, Z.~Lu, J.~Chen, Y.~Li, M.~Zhang, T.~Bai, Y.~Fang, L.~Sun, P.~S. Yu, and C.~Shi, ``Graph foundation models: Concepts, opportunities and challenges,'' \emph{IEEE Transactions on Pattern Analysis and Machine Intelligence}, vol.~47, no.~6, pp. 5023--5044, 2025.

\bibitem{yu2025graph2text}
S.~Yu, Y.~Wang, R.~Li, G.~Liu, Y.~Shen, S.~Ji, B.~Li, F.~Han, X.~Zhang, and F.~Xia, ``Graph2text or graph2token: A perspective of large language models for graph learning,'' \emph{arXiv preprint arXiv:2501.01124}, 2025.

\bibitem{chhikara2024few}
G.~Chhikara, A.~Sharma, K.~Ghosh, and A.~Chakraborty, ``Few-shot fairness: Unveiling {LLM}'s potential for fairness-aware classification,'' \emph{arXiv preprint arXiv:2402.18502}, 2024.

\bibitem{xu2023disentangled2}
Z.~Xu, J.~Liu, D.~Cheng, J.~Li, L.~Liu, and K.~Wang, ``Disentangled representation with causal constraints for counterfactual fairness,'' in \emph{Pacific-Asia Conference on Knowledge Discovery and Data Mining}, 2023, pp. 471--482.

\bibitem{yu2025bridging}
L.~Yu, L.~Guo, P.~Kuang, and F.~Zhou, ``Bridging the fairness gap: Enhancing pre-trained models with {LLM}-generated sentences,'' in \emph{Proceedings of the 2025 IEEE International Conference on Acoustics, Speech and Signal Processing}, 2025.

\bibitem{kumar2025detecting}
R.~Kumar, H.~Kumar, and K.~Shalini, ``Detecting and mitigating bias in {LLM}s through knowledge graph-augmented training,'' in \emph{Proceedings of the 2025 International Conference on Artificial Intelligence and Data Engineering}, 2025.

\end{thebibliography}

\vspace{-12pt}
\begin{IEEEbiography}[{\includegraphics[width=1in,height=1.25in,clip,keepaspectratio]{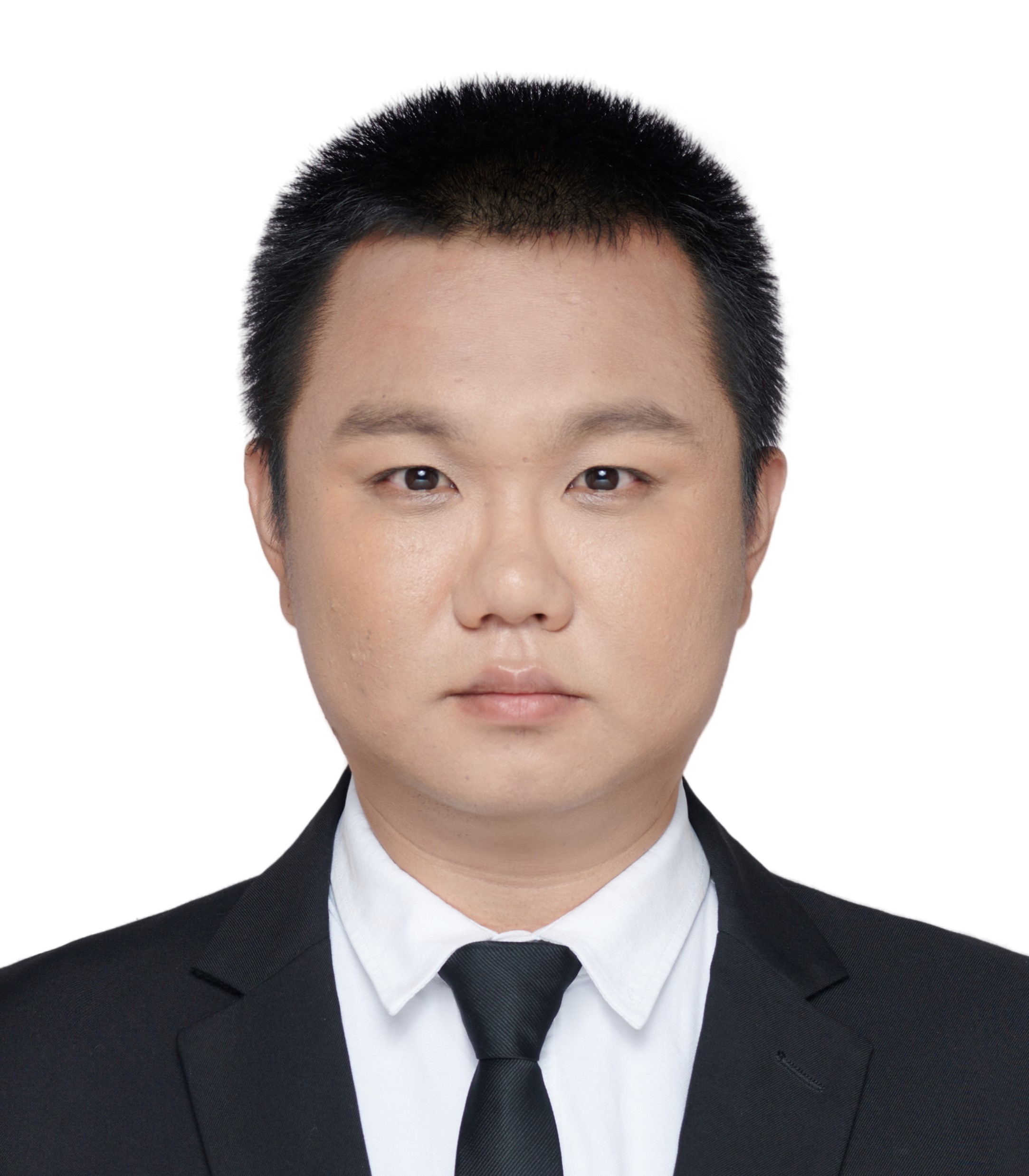}}]{Renqiang Luo}
  received the B.Sc. degree from University of Science and Technology of China, Hefei, China, in 2016, and the M.Sc. degree from University of South Australia, Adelaide, Australia, in 2019. He received a Ph.D. degree in the School of Software, Dalian University of Technology, Dalian, China, in 2024. Dr. Renqiang Luo is currently an Assistant Professor in the Jilin University, Changchun, China. His research interests include graph learning, algorithmic fairness, and trustworthy AI.
\end{IEEEbiography}

\vspace{-12pt}
\begin{IEEEbiography}[{\includegraphics[width=1in,height=1.25in,clip,keepaspectratio]{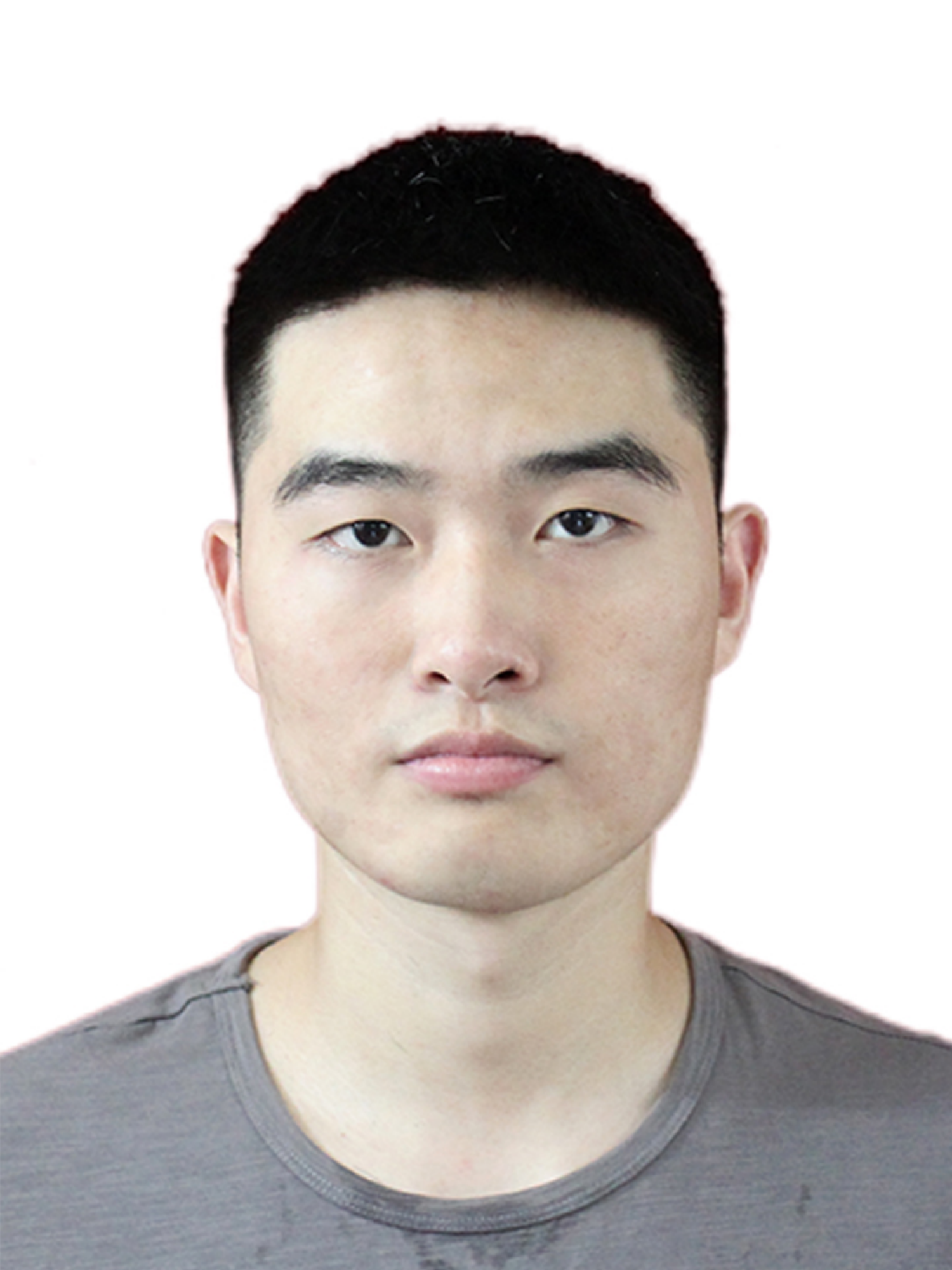}}]{Huafei Huang}
    is currently a PhD student at School of Computer Science and Information Technology, Adelaide University. Before that, he received a BSc. degree from North University of China (NUC), in 2020, and the MSc degree from Dalian University of Technology (DUT), in 2023. His research interests include graph learning and large language models.
\end{IEEEbiography}

\vspace{-12pt}
\begin{IEEEbiography}[{\includegraphics[width=1in,height=1.25in,clip,keepaspectratio]{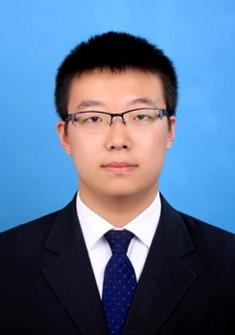}}]{Ziqi Xu} received the M.S. degree in Computing and Innovation from the School of Computer and Mathematical Sciences, The University of Adelaide, Australia, and the Ph.D. degree in Computer Science from the University of South Australia, Australia. 
He is currently a Lecturer in Data Science and Artificial Intelligence with the School of Computing Technologies, RMIT University, Australia. 
His research interests include responsible AI, causal inference, fairness, and explainable machine learning.
\end{IEEEbiography}

\vspace{-12pt}
\begin{IEEEbiography}[{\includegraphics[width=1in,height=1.25in,clip,keepaspectratio]{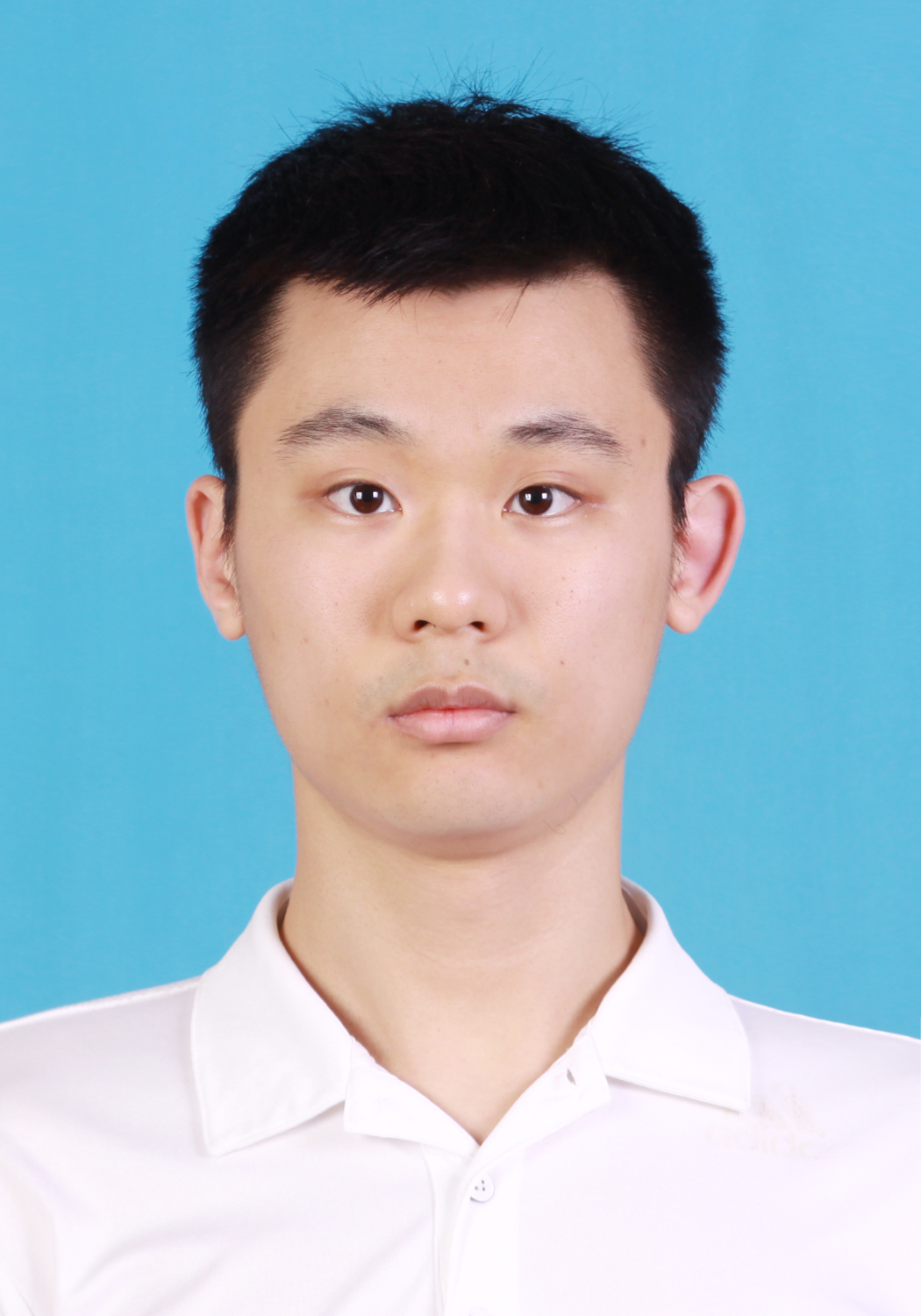}}]{Xikun Zhang} is a Lecturer at the School of Computing Technologies at RMIT University. He received his Ph.D. from the School of Computer Science at the University of Sydney. His research interests span deep graph learning, reasoning with large language models, and biomedical AI. His work has been published in leading conferences and journals, including ICLR, NeurIPS, KDD, ICDM, CVPR, ECCV, TPAMI, and TNNLS.
\end{IEEEbiography}

\vspace{-12pt}
\begin{IEEEbiography}[{\includegraphics[width=1in,height=1.25in,clip,keepaspectratio]{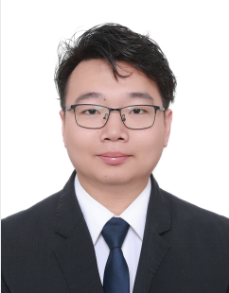}}]{Enyan Dai} is a tenure-track assistant professor at the Hong Kong University of Science and Technology (Guangzhou). 
He received his Ph.D. degree from the Pennsylvania State University.  
Before that he received his master and bachelor degree from KU Leuven and University of Science and Technology of China, respectively. 
His research interests lie in trustworthy AI and its real-world applications. 
For the trustworthy AI, he has accomplished impactful works in aspects of fairness, robustness, privacy and explainability.
\end{IEEEbiography}
 
\vspace{-12pt}
\begin{IEEEbiography}[{\includegraphics[width=1in,height=1.25in,clip,keepaspectratio]{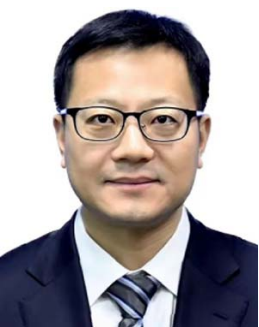}}]{Bo Yang}
is currently the director of the Key Laboratory of Symbolic Computation and Knowledge Engineering of the Ministry of Education, Jilin University, Changchun, China. 
His research interests include data mining, machine learning, knowledge engineering, and complex/social network modeling and analysis. 
He has published more than 120 articles on international journals, including TPAMI, TKDE, TNNLS, and international conferences, including ICLR, NeurIPS, IJCAI, AAAI, WWW, ICDM, and COLING.
\end{IEEEbiography}

\vspace{-12pt}
\begin{IEEEbiography}[{\includegraphics[width=1in,height=1.25in,clip,keepaspectratio]{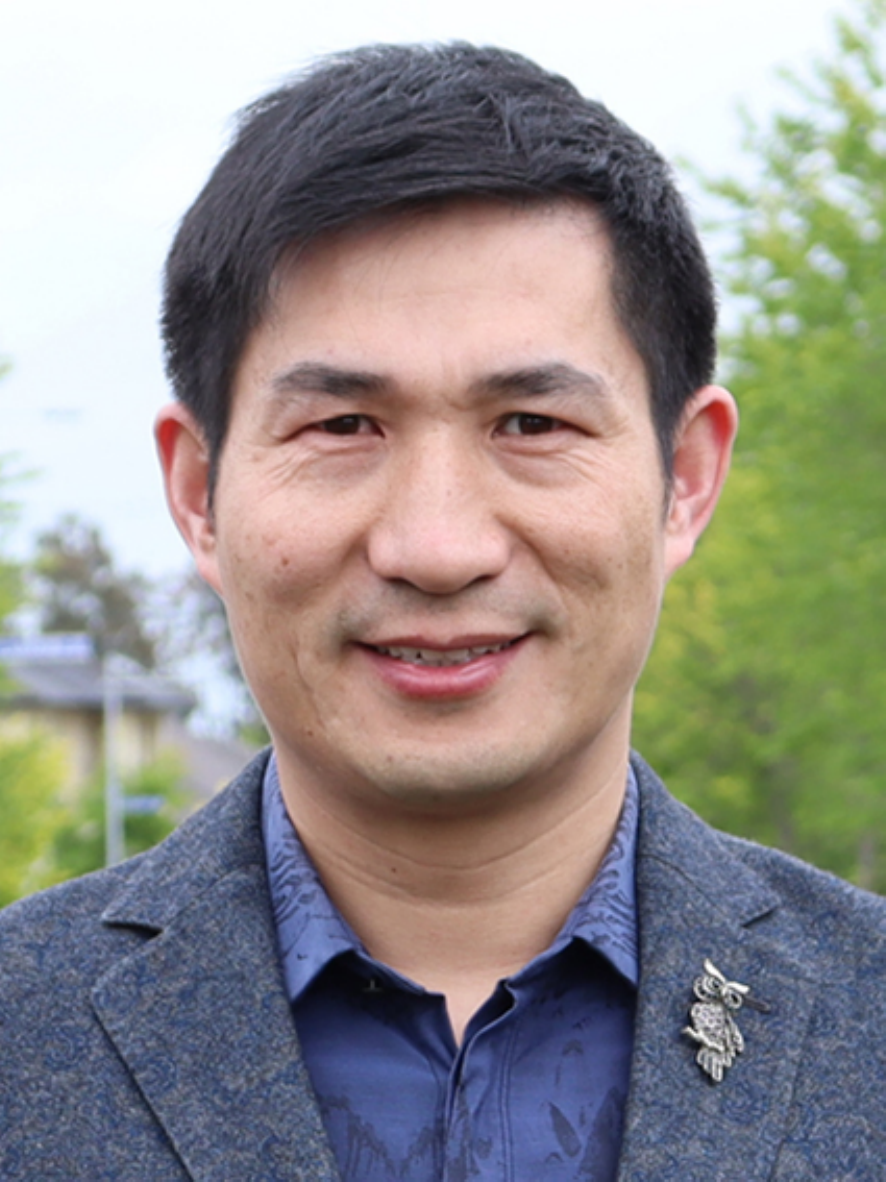}}]{Feng Xia}
(Fellow, IEEE) received the BSc and PhD degrees from Zhejiang University, Hangzhou, China. He is a Professor in School of Computing Technologies, RMIT University, Australia. Recognized as a Clarivate Highly Cited Researcher and a ScholarGPS Highly Ranked Scholar, Dr. Xia has published over 400 scientific papers. His work is featured in top-tier journals and conferences. Dr. Xia has extensive editorial and organizational experience, having served as an Associate or Guest Editor for over 20 journals and in various Chair roles for more than 30 conferences. His contributions and leadership have been recognized by prestigious awards. He has delivered numerous keynote speeches and invited talks at international venues worldwide. He is the Chair of IEEE Task Force on Learning for Graphs. His research interests include artificial intelligence, graph learning, brain, robotics, and digital health. He is a Fellow of the IEEE.
\end{IEEEbiography}

\vfill
\end{document}